\documentclass[preprint,12pt]{elsarticle}

\usepackage{amssymb}

\usepackage{amsmath}
\usepackage{multirow}
\usepackage[table,xcdraw]{xcolor}
\usepackage{float}
\usepackage{adjustbox}

\usepackage{caption} 
\begin{document}

\begin{frontmatter}



\title{Intra-domain and cross-domain transfer learning for time series data – How transferable are the features?}


\author[inst_riteh]{Otović Erik}
\author[inst_riteh]{Njirjak Marko}
\author[inst_ingv,inst_science_roma]{Jozinović Dario}
\author[inst_riteh,inst_airi]{Mauša Goran}
\author[inst_ingv]{Michelini Alberto}
\author[inst_riteh,inst_airi]{Štajduhar Ivan\corref{corresponding_author}}

\cortext[corresponding_author]{Corresponding author}

\affiliation[inst_riteh]{organization={Department of Computer Engineering, Faculty of Engineering, University of Rijeka},
            city={Rijeka},
            country={Croatia}}

\affiliation[inst_ingv]{organization={Istituto Nazionale di Geofisica e Vulcanologia},
            city={Rome},
            country={Italy}}
            
\affiliation[inst_science_roma]{organization={Department of Science, Roma Tre University},
            city={Rome},
            country={Italy}}

\affiliation[inst_airi]{organization={Center for Artificial Intelligence and Cybersecurity, University of Rijeka},
            city={Rijeka},
            country={Croatia}}

\begin{abstract}
In practice, it is very demanding and sometimes impossible to collect datasets of tagged data large enough to successfully train a machine learning model, and one possible solution to this problem is transfer learning. This study aims to assess how transferable are the features between different domains of time series data and under which conditions. The effects of transfer learning are observed in terms of predictive performance of the models and their convergence rate during training. In our experiment, we use reduced data sets of 1,500 and 9,000 data instances to mimic real world conditions. Using the same scaled-down datasets, we trained two sets of machine learning models: those that were trained with transfer learning and those that were trained from scratch. Four machine learning models were used for the experiment. Transfer of knowledge was performed within the same domain of application (seismology), as well as between mutually different domains of application (seismology, speech, medicine, finance). We observe the predictive performance of the models and the convergence rate during the training. In order to confirm the validity of the obtained results, we repeated the experiments seven times and applied statistical tests to confirm the significance of the results. The general conclusion of our study is that transfer learning is very likely to either increase or not negatively affect the predictive performance of the model or its convergence rate. The collected data is analysed in more details to determine which source and target domains are compatible for transfer of knowledge. We also analyse the effect of target dataset size and the selection of model and its hyperparameters on the effects of transfer learning.
\end{abstract}

\begin{graphicalabstract}
\includegraphics{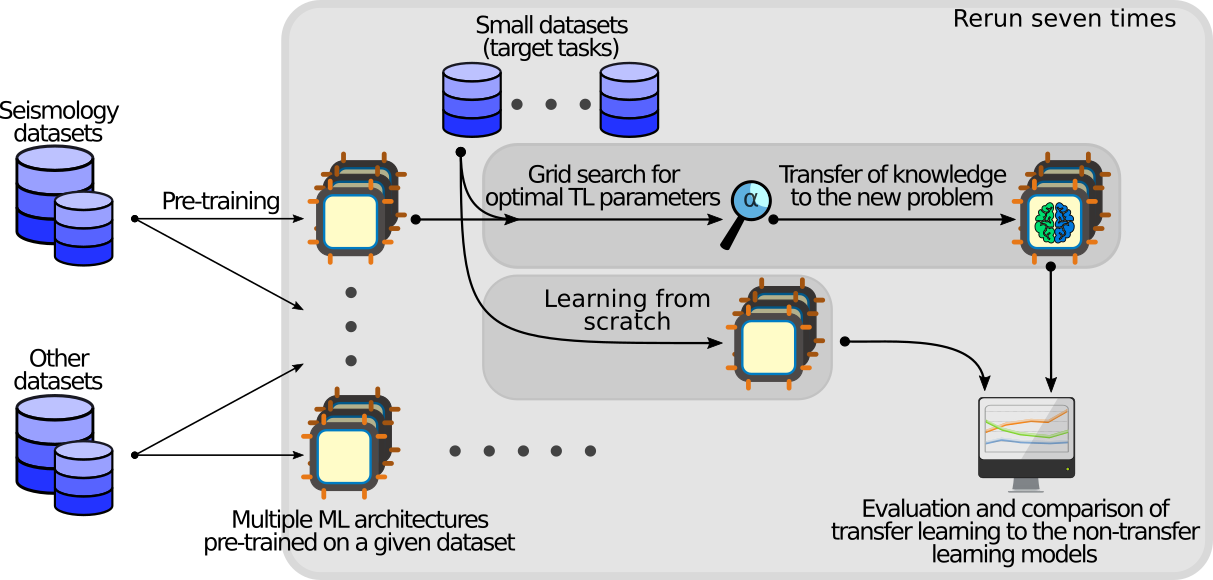}
\end{graphicalabstract}

\begin{highlights}
\item Transfer learning is very likely to either result in positive or non-negative effects
\item The hyperparameters for optimal transfer learning depend a lot on the chosen model
\item A model pre-trained on unrelated task can be better than a randomly initialized model
\end{highlights}


\end{frontmatter}


\section{Introduction}
\label{instroduction}
\label{sec:sample1}
Over the last few years, deep learning techniques have started to become increasingly popular, introducing new and exciting challenges. One of the main challenges and obstacles in the training of deep neural networks is the need for datasets containing sufficient quantities of training instances. Creating such datasets is generally time consuming, which can lead to a slowdown in the application of deep learning in some fields. For example, this may be the case when it is difficult to collect additional data instances because the observed phenomenon is very rare or labelling instances for supervised learning is time consuming because it needs to be done manually. Transfer learning (TL) is one of the possible approaches to combat these problems.

TL allows a machine learning (ML) model trained to solve one problem to be adapted, or fine-tuned, to solve another problem. In this way, part of the knowledge that the model possesses from the first task is used to solve the second task. Knowledge transfer reduces the required number of training instances to solve another task in comparison to training from randomly-initialised models, shortens training time and yields better accuracy. One of the domains in which this approach has proven useful is image classification. There are several state-of-the-art models (like VGG or Inception) pre-trained on large datasets of images that can be fine-tuned to solve some other problems using a significantly smaller dataset and in a much shorter time (see review\citep{6847217}). In this context, TL has enabled the application of these architectures to problems where they could not otherwise be (successfully) applied due to the small amount of training data or due to computationally expensive or lengthy computational operations when training models using large datasets.

However, over the last several years, research has been published, reporting the application of TL for time series (TS) classification and prediction. However, the overall number of reported work is still very limited. Papers on this topic can be divided into two categories. Papers in the first category tend to explain a particular situation in which TL was employed. Such papers can give the reader an impression of good practices in knowledge transfer. Most of such papers are focused on medicine like. For example, in \citep{lin2017improving} TL was used to transfer knowledge about EEG signals of one person to the case of another person for an emotion classification task. In \citep{wen2019time}, the authors presented a TS anomaly detection method that can improve the performance of automated monitoring systems (e.g. in hospital treatment). There are also examples of TS forecasting improved by TL. In \citep{HU201683}, the authors try to predict short-term speed of the wind on a new location by transferring knowledge about wind speed from data-rich locations. In \citep{jozinovic2021transfer}, the authors explore the use of TL for earthquake ground motion prediction using multi-station seismic TS by transferring knowledge from two different seismological datasets with models trained for the same problem or models trained for a different (seismological) problem.

The works from the second category tend to advance and popularize TL for TS data in the same way it was popularized for image classification. These papers are proposing new pretrained models and TL frameworks for TS data. In~\citep{kashiparekh2019convtimenet}, the authors proposed a ConvTimeNet model that is pre-trained and validated on a UCR dataset. In~\citep{ye2018novel}, the authors analyse TS data whose properties are varying over time, and research how to utilize knowledge gained earlier in the time once the properties of series change. 

In~\citep{8621990}, the authors have studied TL for TS data classification using a fully convolutional neural network. They used datasets from several domains obtained from the UCR archive, and tested pairs of datasets - one as the source dataset and another as the target dataset. However, most of these datasets are very small and this is a potential problem because the quality of the pre-trained model, and consequently the benefit of TL, depends on how well the pre-trained model was trained.

All these studies were limited by some parameter: either to one architecture, or one domain of application, or just one dataset, etc., which is why we cannot broadly look at the effects and draw some general conclusions about TL for TS data. To the best of our knowledge, there is no study that has systematically observed the impact of TL in the domain of TS in a broader sense. 

This paper seeks to fill this gap by doing a series of experiments. To get a broader picture of TL, the conducted experiments and tests are focused on the following ideas:

\begin{enumerate}
    \item It is reasonable to expect that knowledge transfer between related domains could be more beneficial than knowledge transfer between unrelated domains. For this reason, this paper tests knowledge transfer within the same domain (intra-domain TL), and knowledge transfer between different domains (cross-domain TL). In the case of intra-domain TL, we test the transfer of knowledge within seismology (different signal characteristics), and in the case of cross-domain TL we test the transfer of knowledge between the domains of seismology, sound, medicine and finance. The reason why cross-domain TL is interesting is that all TS data can be decomposed into a linear combination of sine and cosine waves, and in this way the TS data from different domains are related to some extent. This suggests that there is common knowledge that could benefit the training process of the ML models when applied to two domains that seemingly, in the real world, are not related at all.
    
    \item We use multiple datasets to perform experiments related to intra-domain and cross-domain TL. Some of these datasets are concerning classification tasks, whereas others are dealing with regression tasks. We ensure that datasets are of proper size for pre-training the models. In order to give a better insight into the functioning of TL, we test the transfer of knowledge between all possible pairs of used datasets. While doing this, a different number of channels between source and target datasets poses a problem. We present our solution to this problem before proceeding with TL.
    
    \item As mentioned earlier, learning by transferring knowledge enables the training of models using smaller datasets. However, it is known that the quality of the ML model highly depends on the size of the training set. Therefore, in our experiments, we made two variants of different sizes for each training dataset. This should allow us to see how the effects of TL change with the size of the training set.
    
    \item Not all models are equally suitable for all tasks. Therefore, we expect that some models trained using TL could attain better results than others. Because the aim of this paper is to gain a broader impression of TL, four ML models were used for all experiments: two models were taken from seismology, and the other two models are general-purpose models for TS data. In this way, our conclusions are not tied to any particular model.
    
    \item The success of knowledge transfer depends on the selected hyperparameters that control the training process on the target dataset (i.e. fine-tuning). In our case, we have only one such parameter and it is the learning rate multiplier which regulates how quickly weights of pre-trained layers are modified during fine-tuning. We assume that different values of this parameter will be suitable for different cases. Therefore, we perform a grid-search to determine the optimal value of this parameter for each individual case. In this way, all cases will achieve the best possible results and thus the impact of this parameter on our conclusions is minimized.
    
    \item During model training, the main focus is on getting the best possible performance. However, in other fields it has been observed that pre-trained models converge faster because they already possess some knowledge. Therefore, in this paper we study the performance of the models and the speed of their convergence.
\end{enumerate}

In this sense, this paper is systematic and comprehensive because it seeks to examine all possible combinations of given models, hyperparameters, datasets and training set sizes to eliminate the need to make assumptions that could later bias the results. In this way, we overcome the limitations imposed in other studies and provide a better insight into the effects of TL for TS data.
	
All the source code required to replicate these results has been made public and is freely available for download from the GitHub repository~\footnote{https://github.com/ecokeco/time\_series\_transfer\_learning} along with the obtained results.

\section{Materials and Methods}
\label{materials_and_methods}
In this section, we start by describing the used datasets and models.  Next, we explain the TL process, the obstacles we faced and how we resolved them. Finally, we describe all parts of our experiment: data preparation, model adaptation, training procedure, evaluation metrics and statistical tests. We also provide information about the software and hardware used.

\subsection{Datasets}
\label{datasets}
In this section, we present the TS datasets that were used in our experiments. Three of them are seismic datasets (LOMAX, LEN-DB and STEAD), one is a spoken word dataset (acoustic signals), one is a medical dataset (EMG) and the last one is stock-market prices dataset (S\&P 500). The focus of our study was on TS TL across different domains, but also within the same domain (seismology), which explains the inclusion of several seismic datasets.

\subsubsection{Lomax dataset}
\label{lomax_dataset}
This seismological dataset~\citep{lomax_anthony_2021_5040865} was presented and used in~\citep{lomax2019}. Through this paper we will address it as the LOMAX dataset as they did not name it. This dataset contains 22,046 three-channel (BHZ / N / E) seismograms of global earthquakes at any epicentral distance, collected using the MedNet network of stations in the period from 2010 to 2018. From the same network of stations, 13,009 noise seismograms are collected and provided in the dataset. All recorded seismograms are of duration 50 seconds (1,001 sampling points) and were collected using a sampling rate of 20 Hz. The units of seismograms in this dataset are meters per second.

The earthquake waveforms begin 5 seconds before the first P wave arrives. As a form of quality check, \citep{lomax2019} kept only the earthquake waveforms with signal-to-noise ratio (SNR) greater than 3.0, and noise waveforms with SNR smaller than 1.5.

In \citep{lomax2019}, the dataset was preprocessed in such a way that each instance of seismogram was normalized separately. The maximum value of each waveform signal from all three channels is stored (they named it \textit{stream max}), and this value is later used in the neural network with the aim of improving the results.

As we used the data for the task of earthquake magnitude determination, we decided to filter the test set to contain only the instances that were correctly classified by the convolutional neural network (CNN) model used in the original study (87\% accuracy). We also visually checked the erroneously classified earthquakes (i.e. those that the CNN model classified as seismic noise) and found that most of them were earthquake signals which were buried in the seismic noise, and not easily recognisable as earthquakes. This confirmed our choice of leaving those data out when training the model for earthquake magnitude determination. The dataset also contains some duplicates that we removed. The total size of the resulting dataset containing only the earthquakes was 19,426.

\subsubsection{LEN-DB dataset}
\label{lendb_dataset}
LEN-DB dataset was published in 2020, with the aim of collecting a sufficiently large amount of data for use with ML\citep{magrini_fabrizio_2020_3648232}. It is a global library of local earthquakes created by collecting data from 1,487 broad-band or very broad-band measuring stations deployed around the world. The entire dataset is publicly available online in the form of a single HDF5 file\citep{lendb_zenodo}. The dataset consists of 1,249,411 three-channel seismograms, of which 631,105 seismograms contain earthquakes and 618,306 contain seismic noise. Seismograms containing earthquakes were obtained from 304,878 different earthquakes. Seismograms are 27 seconds long, with a sampling frequency of 20 Hz, which means that one channel of a seismogram contains 540 sampling points. The ground motion in seismograms within this dataset is expressed in meters per second. We kept only those seismograms containing earthquakes because we are doing an earthquake magnitude determination task.

In their paper, the authors also presented a simple example of how to use the dataset for earthquake detection. The model used in the example was a variant of the model used in \citep{lomax2019}. The input data in their paper was normalised and a \textit{stream max} value was determined as was done in \citep{lomax2019}.

\subsubsection{STanford EArtquake Dataset}
\label{stead_dataset}
Stanford EArtquake Dataset (or STEAD for short) is a database of seismograms of local earthquakes collected from stations around the world\citep{8871127}. Thus, this dataset deals with the same phenomena as LEN-DB. The entire dataset is publicly available as an HDF5 file. The dataset file downloaded from the official Github repository is named stead\_waveforms\_11\_13\_19.hdf5. The file was downloaded on 2020-05-03, and the Github repository has been updated several times since then. Using the repository history it is possible to see the repository as it looked on the day we downloaded the file and download the same file we did.

This dataset consists of 1,137,793 three-channel seismograms, of which 1,031,908 represent earthquakes and 105,885 represent noise. All seismograms have a sampling frequency of 100 Hz, and time duration of 60 seconds, giving 6,000 sampling points per seismogram channel. The units of seismograms are counts, which are dependent on the transfer function of the instrument recording the waveform. Because of that, two seismograms of the same ground motion recorded using different instruments may result in differing signal amplitudes. Therefore, seismograms collected by different instruments are not directly comparable. 

\subsubsection{Speech commands dataset}
\label{speech_dataset}
Speech Commands is a dataset compiled and published by Google Brain\citep{warden2018speech}. Through this paper we will address it with the abbreviation SPEECH. This dataset consists of approximately 105,000 WAV files that contain the sound of the recorded word, and the sampling frequency used is 16 kHz. All recordings of spoken words are classified into one of 35 possible classes, and, in addition, there are several longer recordings of noise that contain no speech. We have empirically determined that ML models described in section~\ref{ml_models} have the best performance when we resample the sound to 8 kHz. This is especially true of the ConvNetQuake INGV model, which failed to converge on the original recordings having 16 kHz sampling frequency. The solution to this problem is described in more detail in section~\ref{convnet_model}. Each soundtrack lasts about 1 second, and in preprocessing we reduce all soundtracks to the same duration of exactly 1 second (8,000 sampling points). We do this by supplementing the recordings that last less than 1 second with the noise included in the dataset. In the case of recordings lasting longer than 1 second, we retain the first second (8,000 sampling points) of the sound and discard the rest.

At the time of writing, there are two versions of the dataset, and information about them and instructions on where to download them can be found in the corresponding paper \citep{warden2018speech}. In our experiments, we used version v0.02, which contains more different words (i.e. classes) and is more numerous compared to the previous version.

\subsubsection{EMG dataset}
\label{emg_dataset}
We took the EMG (Electromyography) dataset from \citep{donati_elisa_2019_3194792}. Detailed information concerning the dataset can be found in \citep{8584674}. An EMG signal is a biomedical signal that represents the electrical activity produced by muscles when they are stimulated. This dataset was created by recording an EMG signal using the Myo armband. The armband consists of 8 equally spaced non-invasive sensors that record a signal using a sampling rate of 200 Hz. The application of this dataset is in the recognition of hand movements using ML methods.

The dataset consists of two subsets called Pinch and Roshambo. The only difference between these two subsets is in the performed movements. In the case of the Pinch subset, there are four classes that represent pinches between the thumb and index, middle, ring and pinky finger. The Roshambo subset contains three movements (stone, paper, scissors) about which more information can be found in the accompanying paper. The Pinch subset is more numerous and for that reason we have decided to use it exclusively. The movements contained in it were performed by 22 participants. Each participant performed three sessions in which each gesture was performed five times for two seconds. Between the two movements is a period of relaxation lasting one second.

This dataset contains an eight-channel signal. Using a sliding window that moves through the recorded signals, we get samples of equal lengths, while taking care that there is only one type of movement inside the sliding window. We started with the settings used in \citep{Lobov_2018}, that is, we set the length of the sliding window to 200 ms and moved it by 100 ms in each iteration. However, we found empirically that our models achieve significantly better results when we use 400 ms (80 sampling points) long sliding windows, and a 50 ms step. In the end, this method produced a dataset that consists of 32,438 instances.

In \citep{7303979}, the authors have shown that it is possible to achieve good results utilising a smaller number of channels. Therefore, we retained only three channels due to the fact that all other datasets are single-channel or three-channel. Different number of channels between datasets poses a problem in TL. In this way, only the problems of TL between single-channel and three-channel cases must be handled. Our solution to this problem will be further explained in section~\ref{transfer_learning}.

Preserved channels have indices 0, 2 and 5. Since the sensors are placed in a circle around the arm, we selected these channels so that the distance between the corresponding sensors is approximately equal.

\subsubsection{S\&P 500 dataset}
\label{sp_dataset}
Reviewing existing papers in the field of stock price forecasting, we found that it is a common choice to use the S\&P 500 (Standard \& Poor’s 500) dataset \citep{8666592}\citep{8263105}.  Throughout this paper we will address it with the abbreviation S\&P 500. This dataset is publicly available through Yahoo finance (link/referenca) and can be downloaded for any time period. In\citep{8666592}, the authors argue that S\&P 500 is more stable than individual company stock prices, which presents a better potential for the predictive model. For the purposes of this paper, we have downloaded the data from 30. December 1927 to 26. November 2018 (inclusive).

The data is contained in a single CSV file which contains the following information: Date, Open, High, Low, Close and Adjusted Close prices and Volume on the specific date. Looking at the dates, it can be noticed that records are missing for specific days. This is because on those days the stock market was closed. Since on those days there could be no change in the prices, we treat the data as a continuous TS without missing values.

We took the data extraction method and the stock price prediction experiment from \citep{8666592}. First, we extract only the Close price from the downloaded data, and then we divide this time sequence into smaller parts using a sliding window. The sliding window comprises a consecutive sequence of 50 closing prices, and in each step it advances by a single record starting from the oldest date to the most recent one. The result of this operation is a set of 22,681 TS of length 50 that represent a closing price in 50 consecutive working days. Also, with each obtained series of 50 records we associate the closing price that was valid on the 51st day. The goal of the ML model is to predict the price on the 51st days based on the past 50 consecutive records.

Because these are TS data, we paid special attention to the pre-processing to avoid data leakage between training, validation and test sets. We assign the records into training, validation and test sets in such a way that starting from the oldest to the most recent date we take 70\% of the records for the training set, the next 15\% for the validation set and the last 15\% for the test set. Each set is created by a sliding window over its assigned records. In this way, the model cannot come into contact with the test data during the training in any way, and test data is left for the very end to evaluate the performance of the model, thus ensuring that no data leakage occurs between the sets. Once the data were extracted, they were processed as is described in~\ref{data_preprocessing}.

An overview of the previously described datasets is given in Table~\ref{tab:datasets_overview}.
\begin{table}
\caption{An overview of the chosen datasets and their characteristics.}
\label{tab:datasets_overview}
\begin{adjustbox}{max width=\columnwidth,nofloat=table,vspace=\medskipamount,center}
\begin{tabular}{|l|l|l|l|l|l|l|l|}
\hline
\textbf{Dataset} & \textbf{Task type} & \textbf{Domain} & \textbf{Size} & \textbf{Channels} & \textbf{\begin{tabular}[c]{@{}l@{}}Sampling\\ frequency\end{tabular}}   & \textbf{\begin{tabular}[c]{@{}l@{}}\rule{0pt}{5.5mm}Sampling\\ points per\\ channel\rule[-2.5mm]{0pt}{2.5mm}\end{tabular}} & \textbf{\begin{tabular}[c]{@{}l@{}}Waveform\\ duration\end{tabular}} \\ \hline
\rule{0pt}{5.5mm}LOMAX\citep{lomax_anthony_2021_5040865} \rule[-2.5mm]{0pt}{2.5mm}     & Regression         & Seismology      & 19,426        & 3                 & 20 Hz                                                                   & 1,001                                                                            & 50 seconds                                                           \\ \hline
\rule{0pt}{5.5mm}LEN-DB\citep{magrini_fabrizio_2020_3648232} \rule[-2.5mm]{0pt}{2.5mm}    & Regression         & Seismology      & 629,096       & 3                 & 20 Hz                                                                   & 540                                                                              & 27 seconds                                                           \\ \hline
\rule{0pt}{5.5mm}STEAD\citep{8871127} \rule[-2.5mm]{0pt}{2.5mm}     & Regression         & Seismology      & 1,031,908     & 3                 & 100 Hz                                                                  & 6,000                                                                            & 60 seconds                                                           \\ \hline
SPEECH\citep{warden2018speech}    & Classification     & Audio           & 105,829       & 1                 & \begin{tabular}[c]{@{}l@{}}16 kHz\\ (resampled\\ to 8 kHz)\end{tabular} & \begin{tabular}[c]{@{}l@{}}\rule{0pt}{5.5mm}16,000\\ (8,000\\ after\\ resampling)\rule[-2.5mm]{0pt}{2.5mm}\end{tabular}    & 1 second                                                             \\ \hline
\rule{0pt}{5.5mm}EMG\citep{donati_elisa_2019_3194792}\rule[-2.5mm]{0pt}{2.5mm}       & Classification     & Medicine        & 32,438        & 3                 & 200 Hz                                                                  & 80                                                                               & 0.4 seconds                                                          \\ \hline
S\&P 500\citep{8666592,8263105}  & Regression         & Finances        & 22,681        & 1                 & \begin{tabular}[c]{@{}l@{}}\rule{0pt}{5.5mm}Once every\\ working day\rule[-2.5mm]{0pt}{2.5mm}\end{tabular}        & 50                                                                               & 50 days                                                              \\ \hline
\end{tabular}
\end{adjustbox}
\end{table}

\subsection{ML models}
\label{ml_models}
In this section, we briefly describe the models we selected for our experiment. For each model we provide basic information and the work from which it was taken. Hyperparameters not listed here, such as weight initialization methods or dropout rates, were retained as they were in the original work. Therefore, all modifications in the model architecture and modified hyperparameter values are reported and justified in the following paragraphs.

\subsubsection{ConvNetQuake\_INGV}
\label{convnet_model}
\begin{figure}[htbp]
    \centering
    \includegraphics[width=6.95cm, keepaspectratio]{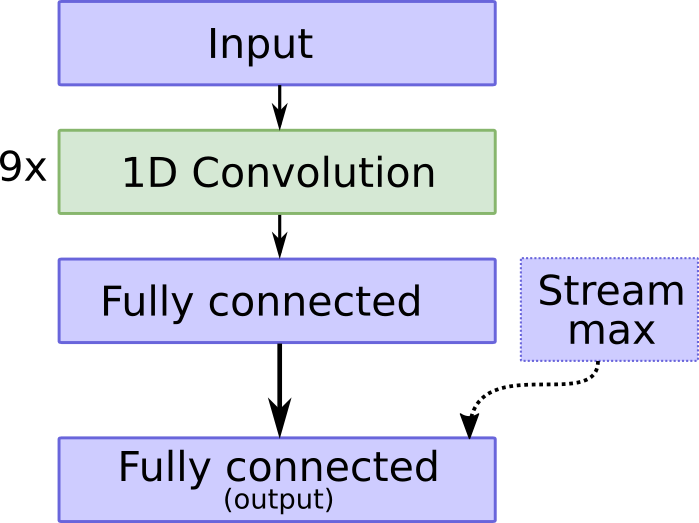}
    \caption{A succinct schematic of ConvNetQuake\_INGV architecture.}
    \label{fig:convnet_model}
\end{figure}

\citep{lomax2019} introduced ConvNetQuake\_INGV (depicted in Figure~\ref{fig:convnet_model}), an adaptation of ConvNetQuake\citep{perol2018}, a CNN model to detect and determine magnitude, location and depth of global earthquakes at any distance (from local to far-teleseismic) over a large range of magnitudes using raw single-station waveforms.

In their experiment, \citep{lomax2019} used the LOMAX dataset for the training and validation. After training, the model was tested on an independent dataset from 2009. Information on the distance, azimuth, depth and magnitude of the earthquake was available for the events. Also, for each stream of input data, a parameter ($stream\_max$) was calculated, which represents the maximum absolute value in a certain waveform across all three channels. The main purpose of this parameter is to aid in CNN magnitude estimation when feeding it the normalized waveform. In their work, the model performed discrete prediction of distance, azimuth, depth, and magnitude into the bins where each bin represented a range of values. To overcome the problem of overfitting, the authors introduced L2 regularization of 0.001 into the convolutional layers.

This architecture has nine 1D convolution layers, whose purpose is to extract features from the input signal. This is followed by two fully connected layers. Outputs of the last convolutional layer are fed to the input of the first fully connected layer along a single $stream\_max$ value. The output layer was modified to match our needs. In our case, the output node contained only a single output neuron for earthquake magnitude estimation. This was necessary because the original model was doing only a discrete prediction of magnitude, while we required a continuous output.

We noticed that this model did not always manage to converge on SPEECH, EMG and S\&P 500 datasets. Therefore, we had to make three variants of this model. Each variant had the same architecture as the original, but with different values for L2 regularization and convolutional layers initializers. For SPEECH dataset, it was sufficient to use Tensorflow’s he\_normal initializer for the convolutional layers and to resample audio files to 8kHz. In the variant intended for the EMG dataset, we changed the initializer to Tensorflow’s glorot\_normal and completely removed the L2 regularization. In order for the model to converge on the S\&P 500 dataset, it was necessary to increase the L2 regularization to 0.7 and use Tensorflow’s he\_normal to initialize the convolution layers.

\subsubsection{MagNet}
\label{magnet_model}
\begin{figure}[htbp]
    \centering
    \includegraphics[width=6.5cm, keepaspectratio]{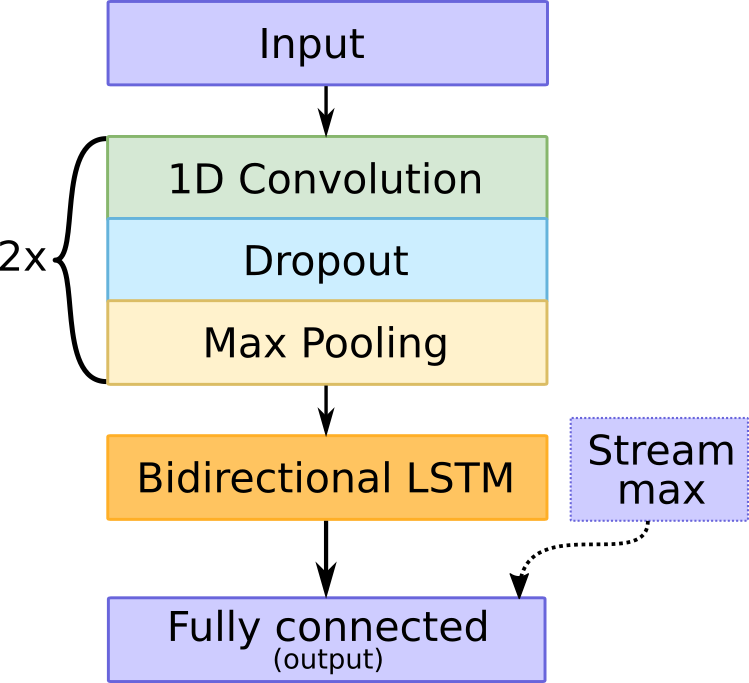}
    \caption{A succinct schematic of MagNet architecture.}
    \label{fig:magnet_model}
\end{figure}

The MagNet model (shown in Figure~\ref{fig:magnet_model}) is a deep learning regression model for end-to-end earthquake magnitude prediction\citep{doi:10.1029/2019GL085976}. The model predicts a magnitude from the raw waveforms of a single station. The authors claim that the model is insensitive to data normalization, hence non-normalized waveforms can be used as inputs. The significance of the MagNet model lies in the fact that it is the first deep-learning approach that has successfully estimated magnitudes from raw-waveform seismic signals obtained from a single station.

This architecture takes a three-channel waveform as its input which is followed by two convolutional layers that do not use any activation function, and their two main roles are to reduce the dimensionality of the data and to extract the features. Each convolutional layer is followed by a dropout layer with a dropout factor of 0.2 and a max pooling layer. The role of the dropout layer is to achieve regularization and each max pooling layer reduces the data 4 times with the aim of shortening the training time. This is followed by one bidirectional LSTM layer with 100 units. \citep{doi:10.1029/2019GL085976} claims that the majority of learning is done in LSTM units and that they are an adequate tool for modeling TS data, such as earthquakes. Ultimately, the output from the network is a single neuron without an activation function (i.e. a linear response neuron). No modifications were done to this architecture because it was perfectly adequate for our experiment.

\subsubsection{MLSTM FCN}
\label{mlstm_model}
\begin{figure}[htbp]
    \centering
    \includegraphics[width=8cm, keepaspectratio]{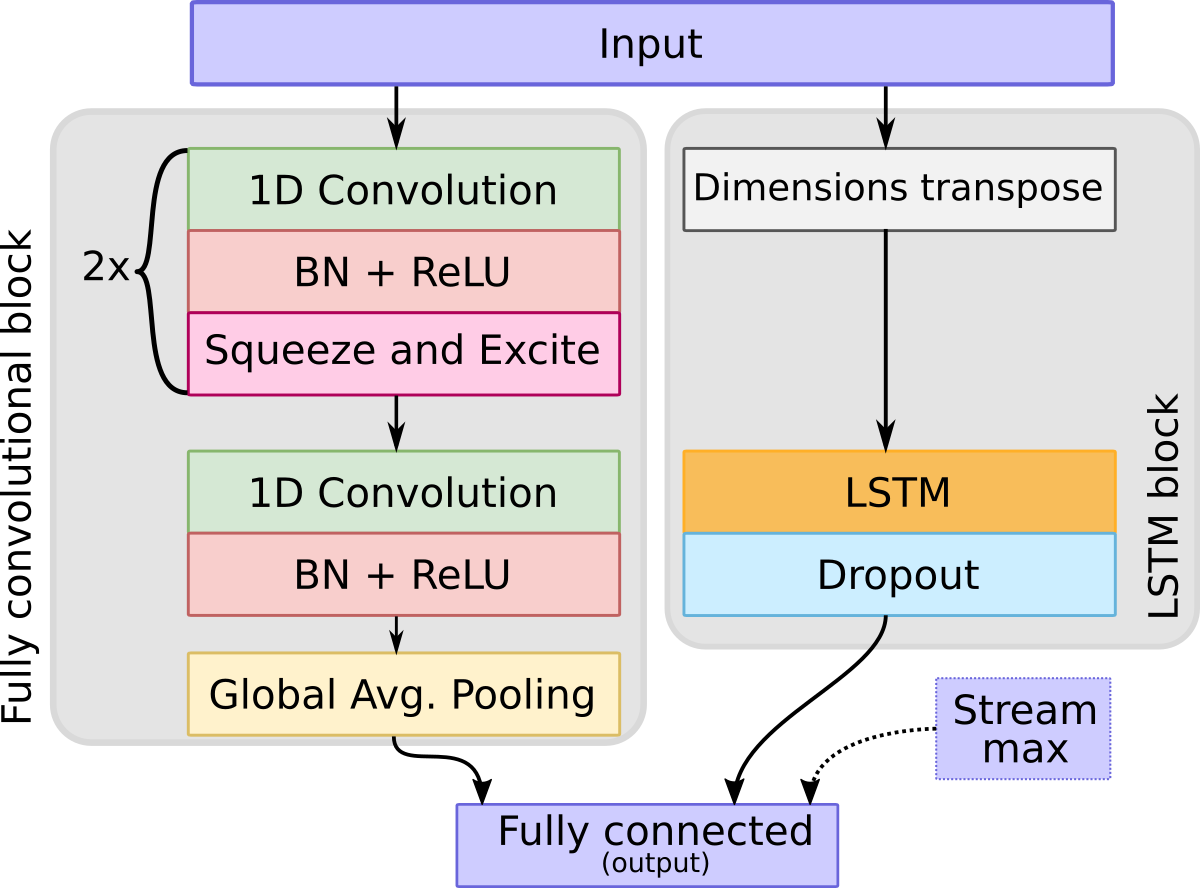}
    \caption{A succinct schematic of MLSTM FCN architecture.}
    \label{fig:mlstm_model}
\end{figure}

In 2019, the Multivariate LSTM Fully Convolutional Network (MLSTM-FCN) architecture (shown in Figure~\ref{fig:mlstm_model}) was proposed, outperforming state-of-the-art models in the classification of complex multivariate TS data\citep{karim2019multivariate}. This architecture is a generalization of the LSTM FCN architecture which is suitable only for univariate signals.

In \citep{karim2019multivariate}, MLSTM FCN was tested on 35 datasets originating from three different sources, and in 28 cases achieved state-of-the-art results. The datasets used were from various domains such as medical care, speech and speaker recognition, and activity recognition. The tasks on which it was tested were very diverse. For example, the smallest number of channels within the task was two, while the largest was 570. In terms of TS length, the shortest length was 15 and the longest was 5,396. It can be seen that all the data we use to conduct the experiment are comparable in size to those on which MLSTM-FCN has been tested. Based on that, we decided to keep the architecture as it is without intervening.

Generally speaking, the model consists of two parallel branches called a fully convolutional block and an LSTM block. The same data are fed to the inputs of both branches, and the outputs of these branches are eventually combined and, based on them, a final prediction is made. A fully convolutional block consists of 3 temporal convolutional blocks. Each temporal convolutional block consists of a convolutional layer, batch normalization followed by a rectified linear unit (ReLU) activation function. Also, the first two temporal blocks contain a “squeeze and excite” block which adaptively recalibrates the input feature maps. The output from the last temporal block is brought to the global average pooling layer and thus the first branch ends. In the second branch, a transpose of temporal dimensions is performed first. This means that the input sequence containing M channels and Q time steps, is converted to a sequence of Q channels and M time steps. This causes the LSTM block to process all data in M steps instead of Q steps. Of course, this only benefits the process if M is less than Q. In their work, the authors show that this significantly shortens the training time, without significantly affecting the accuracy of the model. The temporally-transposed data is fed to a plain LSTM layer followed by a dropout layer to prevent overfitting. A high dropout rate of 80\% is used.

\subsubsection{TCN}
\label{tcn_model}
\begin{figure}[htbp]
    \centering
    \includegraphics[width=6.5cm, keepaspectratio]{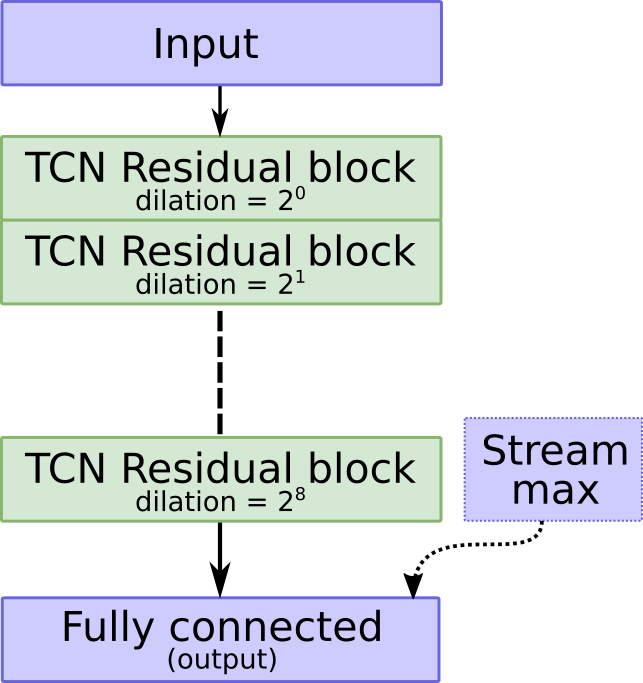}
    \caption{A succinct schematic of TCN architecture.}
    \label{fig:tcn_model}
\end{figure}

In their recent work\citep{bai2018empirical}, the authors presented the results of their research comparing generic convolutional and recurrent sequence modeling networks. In their experiment, they first presented a temporal CNN and then compared its performance on various sequence modeling tasks with the performance of long short-term memory (LSTM), gated recurrent unit (GRU) and vanilla recurrent neural network (RNN) architectures. They explain how TCN represents a simple, yet powerful starting point in sequence modeling and how it achieves better results than generic recurrent architectures.

The main features of the TCN architecture are: 1) the used convolutions are causal, which means that information from the future is not available in the past and 2) the architecture maps the input sequence of any length into an output sequence of the same length as the RNN. This architecture consists of residual blocks that are sequentially connected. Within each residual block is a sequence of layers: Dilated Causal Convolution, weight normalization, ReLU activation, and dropout, and this sequence is repeated twice. The output from the residual block is obtained by summing the inputs to the residual block and the outputs of the last dropout layer. Dilated convolutions enable efficient increase of the receptive field network which, according to the authors, is of great importance for modeling time sequences using TCN. Therefore, the receptive field of TCN can be increased by increasing the number of convolutional layers, using a larger kernel size or a larger dilatation factor. The performance of TCN and generic recurrent architectures was compared on 11 tasks that are mainly used to compare the performance of recurrent networks. When the authors conducted the tests, the used architecture for each task was the same but with different hyperparameters. The following hyperparameters were varied: the size of the convolutional kernel, the number of filters and the depth of the network to control the size of the receptive field.

The choice of hyperparameters for TCN directly affects the receptive field of the model. The receptive field directly depends on the length of the TS data to which the model is applied. As the authors have pointed out, problems may arise when TL is performed because source and destination tasks may have a very different number of time points. This problem can lead to poor TCN performance after TL. Namely, different domains of application may require different lengths of history (i.e. perceptive field) that the model needs to predict the outcome. This is exactly the case in this paper - the S\&P 500 dataset has the fewest time points (only 50), while the SPEECH dataset has the most time points (as many as 8,000). In the original work, the authors tested the model over different lengths of TS data and offered the hyperparameters they used. Hyperparameters used for time sequences that contained 600 time points were taken for the purposes of this paper. These hyperparameters correspond to our "medial" case. In this way, there are three datasets that have fewer than 600 time points (S\&P 500, EMG and LEN-DB) and three datasets that have more than 600 time points (LOMAX, STEAD and SPEECH).

For the sake of reproducibility, these hyperparameters were taken from “The adding problem”, and they are: kernel size 8, number of filters 24, dilation 8, dropout = 0.0 and gradient clip N/A.

ML model obtained by these hyperparameters is depicted in Figure~\ref{fig:tcn_model}.

\subsection{Transfer learning}
\label{transfer_learning}
TL\citep{bozinovski2020reminder} is a technique that seeks to improve the performance of the model on the destination task Tt in the domain Dt using already acquired knowledge from the domain Ds and the task Ts where Ds != Dt or Ts != Tt (or both at the same time). This procedure consists of two phases: in the first phase, the model is trained on the source task and then, in the second phase, it is trained on the destination task. In their research\citep{machine_learning_survey}, the authors present three issues that appear when using TL: "What to transfer", "How to transfer" and "When to transfer".

The "What to transfer" and "How to transfer" questions ask which part of the acquired knowledge from the source domain Ds can be useful in the destination domain Dt and how to transfer that knowledge. In our experiment, we try to take advantage of the fact that convolutional layers learn feature extraction by having earlier layers learn simpler patterns and later layers learn more complex patterns\citep{Tajbakhsh_2016}. Therefore, the model trained in the source domain Ds will retain the convolution filters’ weights before switching to the destination domain Dt in order to preserve the acquired knowledge. How the model will be adjusted during the domain transition and how fine-tuning of convolutional layers is done is described later in the paper.

The "When to transfer" question asks which source domains Ds lead to performance improvements in the destination domain Dt. An opposite effect is also possible in which a pretrained model in the Ds domain performs worse in the Dt domain and this phenomenon is known as negative transfer. The answer to this question is not known for TS domains and is one of the research subjects in this paper. To answer this question, we investigate to which extent the features learned in convolutional layers on one task are useful when transferred to the target task.

When performing TL, two problems occur, which we describe below and explain how we solved them. The first problem is that the same architecture differs in its fully connected layers depending on which dataset it is applied to. Specifically, we use \textit{stream max} (defined in section~\ref{lomax_dataset}) on some datasets while not on others. Also, depending on the type of task there may be more or fewer output neurons. All of the chosen regression problems are dealing with predicting only a single value (earthquake magnitude or stock price), so in their case there is always one output neuron. However, in the case of classification problems, the number of output neurons depends solely on the number of predicted classes. We solved this problem by instantiating a new model that is suitable for the destination task, and convolutional layer filters are copied from the pre-trained model to this model. As already mentioned, this procedure retains knowledge within the convolutional layers, while the weights in the fully connected layers are randomly initialised.

Another problem is the different number of channels between datasets. The problem occurs when we want to fine-tune a model on a target dataset that has a different number of channels than the source dataset on which the model was initially trained. Specifically, the problem occurs in the first convolutional layer whose number of channels depends on the number of channels in the dataset. Because all used datasets have either one or three channels, there are two problematic scenarios. The first one occurs when the source dataset has one channel and the target dataset has three channels. We solved this problem by simply replicating the convolution filters (and their weights) to get three channels in the first convolution layer. This technique is applied when the source data set is SPEECH or S\&P 500 and the target is a three-channel dataset.

Second problematic scenario occurs when the source is a three-channel dataset and the target is a single-channel dataset. More precisely, this occurs when the source dataset is LOMAX, LEN-DB, STEAD or EMG and the target dataset is SPEECH or S\&P 500. We solved this problem by keeping the three-channel convolutional layers as they are, and a single channel in the target dataset is copied to form a three-channel dataset. Notice that signal length difference between source and target datasets is not a problem because signal length does not impact in any way the number of convolutional filters.

As stated earlier, the learned weights of the convolutional layers are preserved and used instead of being randomly initialized before training on the target dataset. These convolutional layers are trained using learning rate conv, while the remaining layers are trained using learning rate . General assumption is that pretrained layers should be trained using different learning rates because they already encode some knowledge. $\alpha_{conv}$ is acquired by multiplying by the factor:

\begin{equation}
    \alpha_{conv} = \omega \cdot \alpha
\end{equation}

For the  value, we selected 10 different values to examine. This is because we have noticed that not all models reach the maximum improvement at the same value of . Therefore, we performed a grid search, i.e. we ran the fine-tuning process 10 times, but each time using a different value of . For the values of , we used the values from the set {0.01, 0.05, 0.1, 0.25, 0.5, 0.75, 1.0, 1.25, 1.5, 2.0}. Although it is common practice to fine-tune pre-trained layers using a lower learning rate, through empirical experimentation (section~\ref{influence_of_models_on_tl}) we discovered that using larger learning rates can also be beneficial in some cases.

\subsection{Experimental setup and evaluation metrics}
\label{experimental_setup_and_evaluation}
In this section, we describe how we preprocessed the data, what modifications we made to the models, what hyperparameter values we used to perform the experiment, the workflow of the experiment and the metrics we used to evaluate the experiment.

\subsubsection{Data pre-processing}
\label{data_preprocessing}
Pre-processing and normalization of data is an important step because, depending on the chosen method, better or worse model predictive accuracy can be attained, and the time required for training can be either shortened or extended \citep{NAWI201332}. The method we applied to all the datasets is described below, regardless of which method was used in the original papers. Before we could apply the chosen method, we had to randomly divide the datasets into training, validation and test sets in percentages of 70\%, 15\% and 15\%, respectively. We introduce the following notation to explain the data preparation process: $S_{train}$ for the training set, $S_{val}$ for the validation set, and $S_{test}$ for the test set. Each of these subsets consists of a number of data instances, and each data instance consists of the pair $(X_i, Y_i)$ where $X_i$ represents the observations that are the input to the ML model, while $Y_i$ represents the ground truth value that the model is trying to predict. An observation $X_i$ is a matrix of dimensions $CxN$, where $C$ represents the number of input channels, and $N$ the number of sampling points recorded over time. $i-th$ data instance that belongs to the training set will be referred to as $(X_{train,i}, Y_{train,i})$. Data instances in the validation and test sets will be called analogously.

$S_{train}$ is the first one being processed, and as the first step we find the minimum and maximum value for each input matrix $X_{train,i}$, and denote them by $m_i$ and $M_i$, respectively. From these values, we can determine the corresponding $stream\_max$ value for the $i-th$ data instance by $stream\_max=max(|m|,|M|)$. Then, using the $min-max$ method, all values $x_{j,k}$ of the matrix $X_{train,i}$ are scaled to the range [0,1] using the expression: 

\begin{equation}
    x_{j,k}'=\frac{x_{j,k} - m}{M - m}
\end{equation}

The last step in the preparation of the training set is to center each element of the matrix $X_{train,i}$ around zero by observing each position within the matrix separately through all training instances. An element at position $(j,k)$ would be centered by subtracting from its value the average value of all elements at that position within the matrix $X_{train,i}$. In doing so, $j$ is the index of the channel currently being observed $(j \in [1,C])$, and $k$ is the time point index in the TS $(k \in [1,N])$. The average value for each pair $(j,k)$ is obtained by dividing the sum of the values at position $(j,k)$ across all $X_{train,i}$ by the number of data instances in the training set.

This process can also be performed using matrix operations which improves performance and speeds up data preprocessing. First, we define a matrix of dimensions CxN which will contain the average value for each position (j,k), and we call it mask. Using matrix operations this can be easily obtained as:

\begin{equation}
    mask = \frac{ \sum_{i=1}^{|S_{train}|} X_{train,i} }{|S_{train}|}
\end{equation}

Then centering each value around zero can also be performed using the matrix operations $X'_{train,i} = X_{train,i} - mask$ for $i \in [1, |S_{train}|]$.

Pre-processing of $S_{val}$ and $S_{test}$ is the same as pre-processing of $S_{train}$, except that the new mask is not calculated - instead, the one calculated during the pre-processing of $S_{train}$ is used. Therefore, we first determine $stream\_max$ for each instance, then scale each instance using the min-max method to the range [0,1], and finally subtract the $mask$ from each instance in the validation and test sets. Such data preparation methods (normalisation and standardisation) are necessary for faster and better convergence (unless models are known to be insensitive to the statistical properties of the input values).

We use reduced variants of the mentioned datasets as the target datasets during TL, in order to simulate target datasets that are small in size. This will allow us to simulate a real life situation when it is not possible to acquire a big enough dataset and to investigate how the size of the target training set affects the model. For each mentioned dataset we make two reduced variants which we call \textit{1k5} and \textit{9k}. We obtain these variants by reducing only the training sets to 1,500 and 9,000 instances, while keeping the test and validation sets as they are. When reducing the training set, we randomly choose which data instances will be contained in the reduced dataset, while preserving the underlying distribution of classes (i.e. using stratification for classification datasets).

Because no resampling of the TS data is done in the preparation step, the models will perceive all the data as if they were sampled using the same sampling frequency. This also affects the models understanding of the frequency content of signals from different domains, as the models perception of the content is affected by the sampling rate of the signal.

\subsubsection{Model adaptation}
\label{model_adaptation}
There are four datasets in our experiment that are tied to a regression problem (LOMAX, LEN-DB, STEAD, S\&P 500), and two datasets tied to a classification problem (SPEECH and EMG). Therefore, we must adapt each model to each task separately. We adjust the model by changing only the last fully connected layer. In the case of regression problems, the last layer contains a single neuron because only a single value is predicted. In general, we could add as many neurons as values predicted in the case of multi-target regression. No activation function is applied to the last fully connected layer. In the case of a classification problem, we add as many neurons to the last layer as there are classes that are predicted using the softmax activation function.

Another modification of the model refers to the \textit{stream max} input. This paper is highly inspired by the work presented in \citep{lomax2019}\citep{mousavi2020}. However, these two papers have opposite views on the use of the \textit{stream max} input. However, this is probably a consequence of using different input data: in \citep{lomax2019}, they use the data in m/s, while the data used in \citep{mousavi2020} do not have instrument response removed (i.e. the data are in counts). Because the information concerning the magnitude of the signal is lost due to the min-max scaling during data preparation, this effectively means that the \textit{stream max} input carries the important information in \citep{lomax2019} and does not carry important information in \citep{mousavi2020}, because one has to know the characteristics of the instrument (i.e. instrument response) to correlate the counts with the ground motion. We decided that all models on one dataset must either have or not have the \textit{stream max} input. This way, all models are given equal inputs which makes their comparison more fair. Therefore, we applied the \textit{stream max} input to all models running over the LOMAX dataset, and we did not apply it to models running over STEAD. For other datasets, we had to conduct experiments to determine whether or not it was better to use the \textit{stream max} input. The results of these experiments are presented later in the Results section.

\subsubsection{Loss functions and training procedure}
\label{loss_fns_and_training_procedure}
The used loss function for regression tasks was Mean Squared Error (MSE) and is shown below. Assume that the performance of the model is measured over N instances, and for each instance the model predicts the value $predicted_i$, and we denote the actual value by $ground\_truth_i$. Then the MSE can be calculated as:

\begin{equation}
    MSE = \frac{1}{n} \sum_{i=1}^{n} |predicted_i - ground\_truth_i|^2
\end{equation}

Categorical Crossentropy (CE) was used in the cases of classification tasks. In the case of a single-label classification in which each sample $i$ must be classified as one of the $C$ classes, the cross entropy for that instance can be calculated as:

\begin{equation}
    CE_i = -ln(p_i)
\end{equation}

where $p_i$ denotes the probability outputted by the last (softmax) layer for the correct class. Since training is done in batches, the single $CE$ value is obtained by averaging computed cross entropies for instances within a single batch.

The objective is to minimize the loss function whether it is a regression or a classification task. We perform this by applying the Adam optimizer using a learning rate = 0.001, which decreases if there is no improvement in the validation loss function for four consecutive epochs. The learning rate is reduced by a factor of 0.2, but in such a way that the learning rate is never less than $0.5 \cdot 10^{-6}$. Similarly, we add an early stopping mechanism that suspends the learning process if loss over validation set is not decreased within ten consecutive epochs to avoid overfitting. However, if the training process is not stopped by early stopping earlier, it is suspended after a maximum of 250 epochs. By reviewing the works from which we took the models, we found that all the models converged well before the 250th epoch. Therefore, this number of maximum epochs should give all models enough time to converge.

\subsubsection{Validation metrics}
\label{validation_metrics}
The aim of this paper was to check how TL affects the performance and the convergence of the model in given situations. In this section, we define the metrics that will quantitatively express model performance. For regression and classification tasks, we will define two metrics for a performance score, and the third that shows how quickly the predictive performance improves.

\paragraph{Classification metric}\mbox{}\\
We use a weighted F1 score to measure the success of classification in classification tasks. The reason for this is the different number of instances within individual classes in the SPEECH dataset, which is why it is not possible to compare models fairly using classification accuracy. F1 score represents the harmonic mean of the test’s precision and recall. The model has excellent precision and recall when F1 score equals one, and worst when it is zero. However, this metric is not directly applicable in our cases because the classification problems in this paper contain more than two classes, and F1 score is intended for binary classification problems. Therefore, we first calculate the F1 score for each class separately, in such a way that this class is opposed to all other classes. This approach is known as one-vs-all F1 score. The values obtained are averaged, but so that the weight of each value is proportional to the number of classes for which it is calculated (hence, weighted F1 score). This is necessary to count in the unequal distribution of instances between classes.

\paragraph{Regression metrics}\mbox{}\\
We use the Mean Absolute Error (MAE) metric to measure the performance of regression models. The MAE shows the average absolute error that the model makes. Namely, in all regression problems presented in this paper, it does not matter whether the difference between the predicted and the actual value is positive or negative. The unit of measurement of this metric is the same as the unit of measure in the dataset over which it is applied (e.g. in the case of the S\&P 500 dataset it is dollars, and in the case of the LEN-DB dataset it is the magnitude of the earthquake)\citep{botchkarev2018performance}. Unlike MSE, which gives more weight to large errors and less weight to smaller ones, MAE shows an average error and therefore the percentage difference calculated between MAE of the two models is easier to interpret. MAE is computed as follows:

\begin{equation}
    MAE = \frac{1}{n} \sum_{i=1}^{n} |predicted_i - observed_i|
\end{equation}

\paragraph{Convergence rate}\mbox{}\\
One of the advantages of TL often mentioned in the literature concerns faster convergence of model optimization due to prior knowledge. A naive way to measure this would be through the number of epochs required to complete the training. However, this metric would not be appropriate because the predictive performance of one model could rapidly increase in the first few epochs and then increase minimally through many more epochs which would avoid an early stopping mechanism, while the other model could gradually reach its maximum performance in less epochs and then being stopped by early stopping. By comparing these two models by epochs, one would come to the conclusion that the second model had faster convergence because of a smaller number of epochs, while in fact, the first model had much more rapid convergence but it took longer for the early stopping mechanism to stop the training. Therefore, in the absence of a more appropriate measure, we introduce the convergence rate metric. We will first give an example that shows the intuition behind this metric, and then we will define it for both the classification and the regression task.

Suppose a situation in which we compare two models whose training was automatically stopped by an early stopping mechanism after the same number of epochs. Also, assume that the first model rapidly became better through the first few epochs, while the second model evenly became better throughout the entire training time. In this case, both models trained for the same number of epochs, but the first model has a higher rate of convergence because its predictive performance improved faster.

\begin{figure}[htbp]
    \centering
    \includegraphics[width=13cm, keepaspectratio]{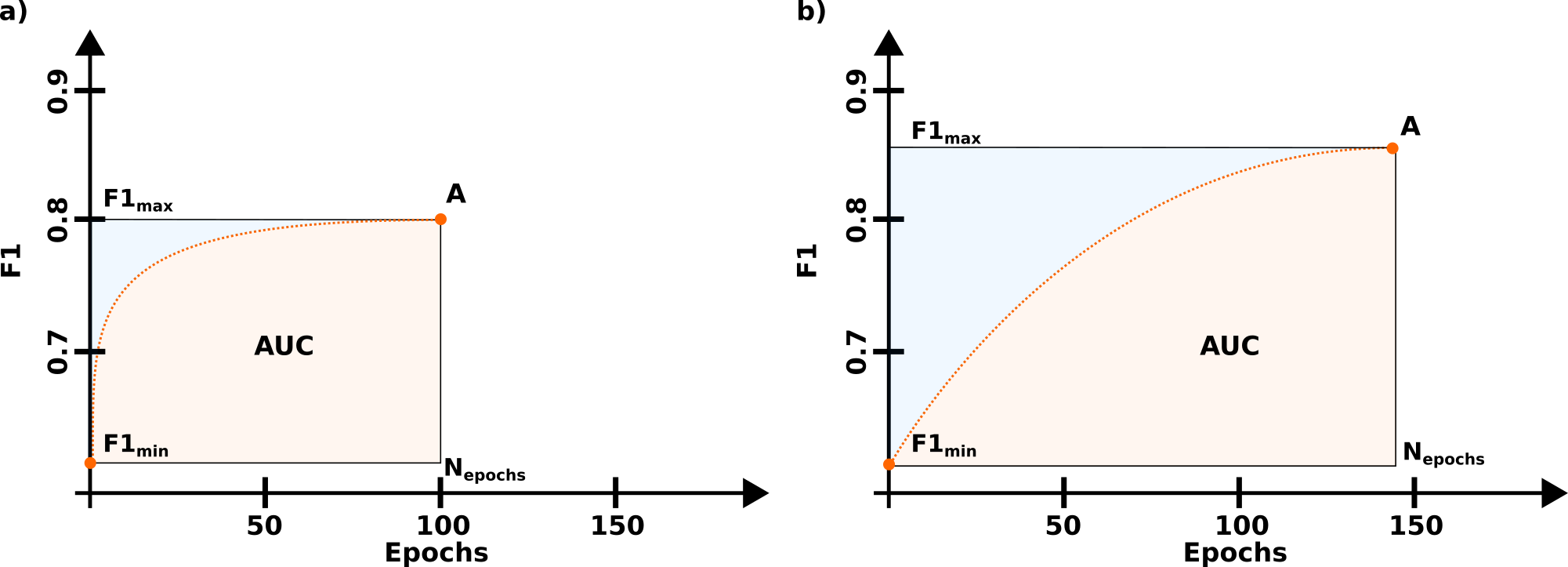}
    \caption{An example of how the convergence rate is computed for the classification tasks.}
    \label{fig:classification_convergence_rate}
\end{figure}

An example of how to calculate the rate of convergence for classification models is shown in Figure~\ref{fig:classification_convergence_rate}. It can be seen that the faster the model gets better, the bigger the AUC area will be in a given rectangle. Naturally, the area of AUC depends on the achieved F1 scores and the number of required epochs, as it can be seen in subfigures \ref{fig:classification_convergence_rate}(a) and \ref{fig:classification_convergence_rate}(b). For this reason, AUC value cannot be directly used to compare the rate of convergence of the two models. In subfigure \ref{fig:classification_convergence_rate}(a), AUC is greater than the one in \ref{fig:classification_convergence_rate}(b), even if the curve in \ref{fig:classification_convergence_rate}(a) shows faster convergence. Because the convergence rate indicates how fast does the model become good (i.e. the shape of the curve), we must eliminate the impact of F1 score and the number of epochs on it. This can be achieved by observing what percentage of area does the AUC occupy in a given rectangle. If we label the total area of a given rectangle as A and the number of epochs as $N_{epochs}$, then the convergence rate can be calculated as:

\begin{equation}
    convergence\_rate = \frac{AUC}{A} = \frac{AUC}{(F1_{max} - F1_{min}) \cdot N_{epochs}}
\end{equation}

The convergence rate takes the value from the range [0, 1] where higher values indicate a faster convergence.

\begin{figure}[htbp]
    \centering
    \includegraphics[width=13cm, keepaspectratio]{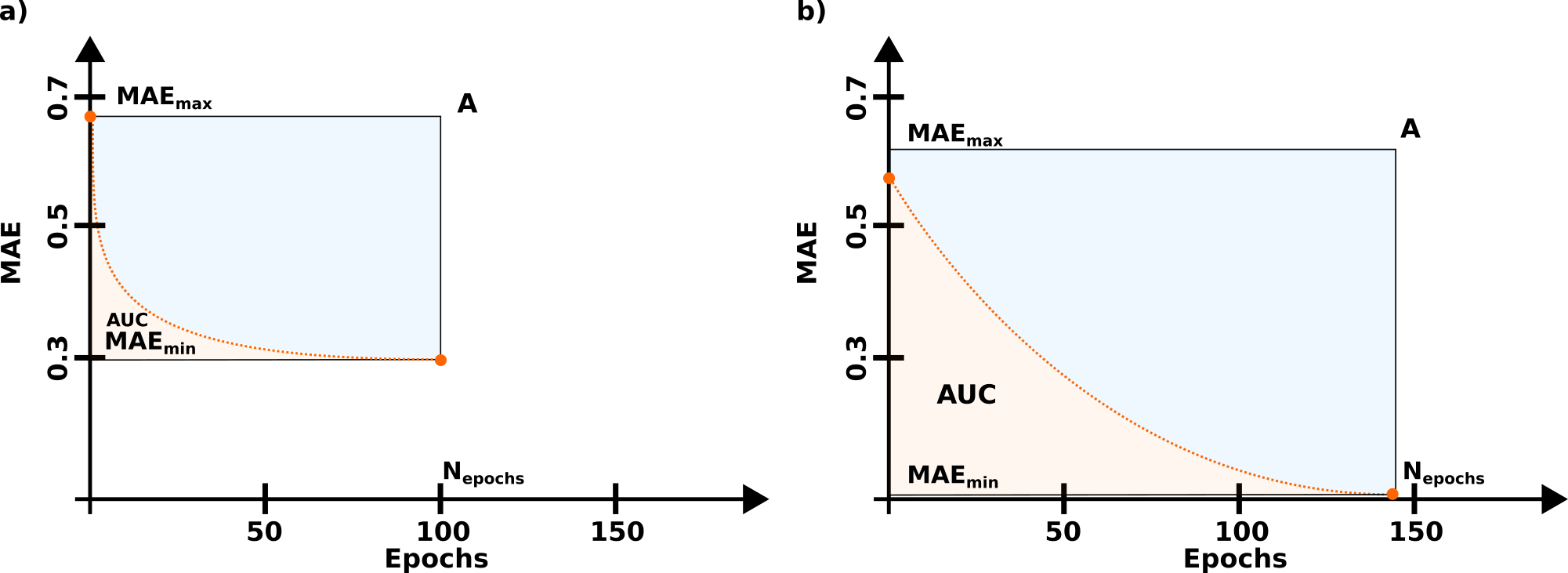}
    \caption{An example of how the convergence rate is computed for the regression tasks.}
    \label{fig:regression_convergence_rate}
\end{figure}

The same can be done for regression models, but unlike classification models where the tendency is to attain a higher F1 score, for regression tasks the goal is to attain a lower MAE. For this reason, the example charts for the regression task shown in Figure~\ref{fig:regression_convergence_rate} are vertically inverted. Another difference is that MAE can take any value in the range $[0, +\infty>$. However, this is not a problem because the convergence rate is representing only the ratio of the two areas. If we label the total area of a given rectangle as A, then we can express the convergence rate for regression models as:

\begin{equation}
    convergence\_rate = \frac{AUC}{A} = \frac{AUC}{(MAE_{max} - MAE_{min}) \cdot N_{epochs}}
\end{equation}

\subsubsection{Experiment workflow}
\label{experimental_workflow}
\begin{figure}[htbp]
    \centering
    \includegraphics[width=13cm, keepaspectratio]{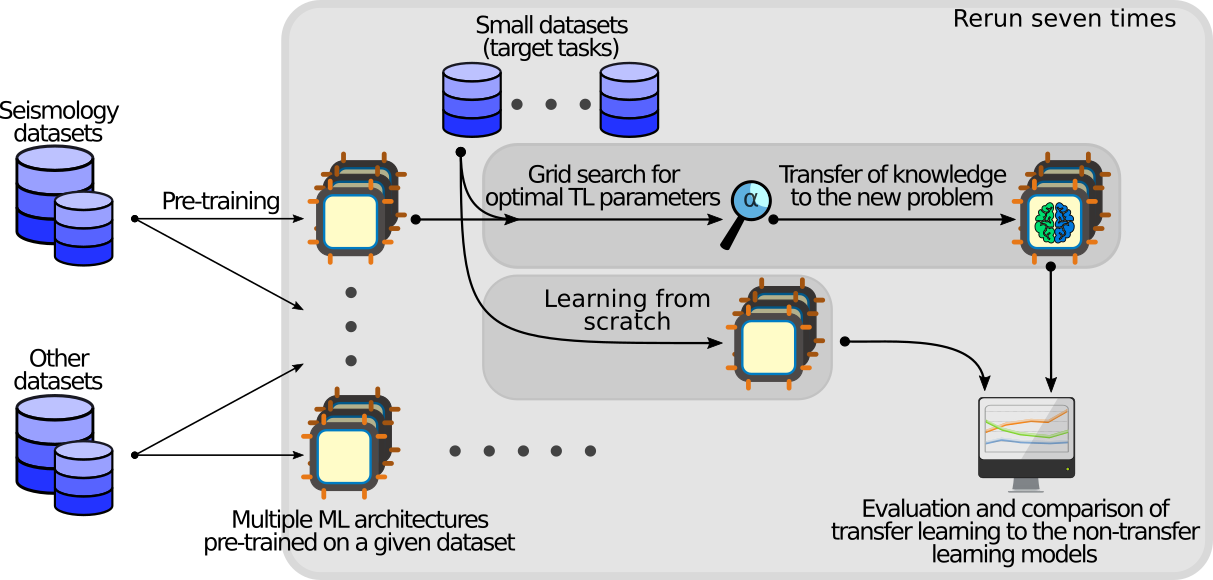}
    \caption{Experiment workflow.}
    \label{fig:pipeline}
\end{figure}

The entire workflow of our experiment is shown in Figure~\ref{fig:pipeline}. Pre-trained models are created by training all four ML model architectures on the six source datasets (three from seismology and three from other domains), which yields 24 pre-trained models in total. It is a common practice when using TL to pre-train a model multiple times over the same dataset and then use the model exhibiting best performance for TL. We take the same approach in our experiment: each pre-trained model was trained from scratch (i.e. from randomly initialised weight values) ten times over the same data and the best one was selected for TL. However, it is still important to manually check if all pre-trained models successfully converged because this can directly affect the quality of TL. We ensure the successful convergence  by checking that the model's performance score is similar to what was reported in the paper that introduced the used dataset.

The second step of the experiment consists of training referent models (non-TL models) from scratch and training TL models by fine-tuning the pre-trained models, and then comparing their performances in terms of the deviation in MAE and weighted F1 scores. The comparison is always done between the pairs of models (e.g. fine-tuned pre-trained MagNet model is always compared to the fully trained MagNet model). To confirm that the observed changes in the MAE and weighted F1 metrics are not coincidental, we repeat the second step seven times (as noted in Figure~\ref{fig:pipeline}) and apply statistical tests to confirm our observations. In each run, a grid search is performed that tries ten different values for hyperparameter (the learning rate multiplier) to find the one that produces the best results when fine-tuning the pre-trained models. This step takes a long time and is the main reason why it was not feasible to rerun the fine-tuning part for more than seven times using the hardware at our disposal.

We have randomly reduced only the training sets to the variants having 1,500 and 9,000 training instances in each run and call them 1k5 and 9k variants, respectively. Validation and test sets remained unchanged which means that the models were validated and tested, using the same data, in each run. Therefore, no data leakage could occur between the training sets and the validation or test sets. TL models are obtained by training (fine-tuning) the pre-trained models from the first step.

We should emphasize that it is not necessary to create pre-trained models from scratch in each run because this would not affect the experiment if they were successfully trained, as we described.

\subsubsection{Statistical tests}
\label{statistical_tests}
The main goal of this paper is to determine whether TL can be considered generally a useful training technique when working with TS data. To do this, TL models are compared to their corresponding referent models. Some of these models are regression models whose performance is measured by MAE, whereas others are classification models whose performance is measured by the weighted F1 score. These two metrics are not commensurable and an appropriate statistical test is needed for this comparison. For this reason, we count wins, losses and ties between models and apply a two-tailed sign test as proposed in \citep{demsar}. This test does not assume commensurability of scores or differences, nor does it assume normal distributions. It is done by examining each pair of the TL model and the corresponding referent model. If the TL model performed better, then it counts as a win for the TL models and as a loss for the referent models (non-TL models). If it performed worse, then it counts as a win for the referent models (non-TL models) and as a loss for the TL models. It can be said that one of the two approaches (TL vs. learning from scratch) is significantly better than the other with significance level of 0.05 if the total number of wins for the considered approach is at least $\frac{N + 1.96\sqrt{N}}{2}$, where $N$ represents the total number of considered cases. If there is no difference between the two methods, then they should have approximately the same number of wins. More details on this test can be found in the referenced paper.

The second goal of this paper is to assess the efficiency of intra-domain and cross-domain TL. This is done by examining each pair of source and target domains. For example, one possible pair for intra-domain TL is picking the STEAD dataset as the source domain, and the LEN-DB (the 1k5 or the 9k variant) dataset as the target domain. Both of these datasets come from seismology, hence intra-domain TL. Similarly, picking the EMG dataset as the source domain and the SPEECH (the 1k5 or the 9k variant) dataset as the target domain would be considered as one possible example of cross-domain TL. Therefore, for each pair of domains a separate statistical test is performed. Measured performance metrics are commensurable because in each case only a single target domain is considered. Hence, a considered metric will either be the MAE or the weighted F1 score, but not a mixture of both. Therefore, a statistically more powerful test can be applied, namely the Wilcoxon signed-ranks test\citep{wilcoxon1945}. It is a nonparametric statistical test for hypothesis testing and is valid for paired data (hence, it is a paired difference test). The null hypothesis assumes that the medial difference between the performance metrics for TL models and referent models is zero. The alternative hypothesis this paper wants to prove is that a significant difference exists between TL models and referent models for a given target dataset. For the sake of understanding, suppose that the source dataset is EMG and the target dataset is LEN-DB (the 1k5 or the 9k variant). This statistical test will check if the TL model that was pre-trained on EMG and later trained on LEN-DB performs significantly better or worse than the referent model that was trained from scratch on LEN-DB (the 1k5 or the 9k variant).

Obviously, Wilcoxon signed-ranks test will be performed multiple times. This can lead to false discoveries and it is a common problem in statistics, known as multiple comparisons problem. Therefore, we apply a well accepted two-stage Benjamini-Krieger-Yekutieli method to control the false discovery rate and compute corrected p values. Corrected p values are then compared to the significance level of 0.05 to either reject the null hypothesis or to keep it.

\subsubsection{Used software and hardware}
\label{used_software_and_hardware}
All program code is written in the Python programming language, and was executed by a Python 3.7.7 interpreter. To train the ML model, we used the popular Tensorflow library\citep{tensorflow} (version 1.14.0) compiled for GPUs. We used the Keras library\citep{keras}, built into Tensorflow (also known as tf.keras), which made the program code simpler and more understandable. Python package keras-tcn (version 3.1.0) was used to instantiate the TCN model and this package is maintained by the authors who introduced the TCN model. We used the keras-lr-multiplier library (version 0.8.0) to be able to apply different learning rates in different layers of the models during TL. All software packages are installed within the virtual Anaconda environment. The GitHub repository contains an Anaconda environment file for automatic creation of a virtual environment that has all the necessary packages installed to run our program and reproduce the results.

The program was run on three Dell EMC PowerEdge C4140 having two Intel Xeon Silver 4114 CPUs and 384 GB of RAM per server. The training was done on twelve NVIDIA Tesla V100 GPU (four per server). At the end, it took about 4.4 TB of disk storage to store all the data and results.

\section{Results and Discussion}
\label{results_and_discussion}
\subsection{Usefulness of stream max input}
\label{usefulness_of_stream_max}
The purpose of exploring the usefulness of the \textit{stream max} input is necessary in order to determine which datasets it should be applied to. For this, each model was trained on each dataset two times: in the first case, the models were trained without the \textit{stream max} input and in the second case with the \textit{stream max} input added. The results obtained in this way are compared in Table~\ref{tab:stream_max_gains}, in which a positive value represents an increase in predictive performance. As can be seen in Table~\ref{tab:stream_max_gains}, the \textit{stream max} input leads to a small improvement in LEN-DB, SPEECH and EMG datasets, and plays a significant role for the S\&P 500 dataset. Therefore, we decided to use the \textit{stream max} input for these datasets. LOMAX and STEAD datasets were left out of this test because their authors have stated in their papers whether \textit{stream max} is useful or not in their cases.

All models achieved a noticeable performance gain on the S\&P 500 dataset when the \textit{stream max} input was added. This is reasonable because without such input, it is impossible for the model to predict the absolute stock price for the next day because the absolute magnitude of the series is lost during data preprocessing.

\begin{table}[H]
\caption{Performance gains when using \textit{stream max}. Positive difference in the case of LEN-DB and S\&P 500 represents a decrease in MAE, while positive difference in the case of SPEECH and EMG represents an increase in the weighted F1 score.}
\label{tab:stream_max_gains}
\begin{tabular}{|l|l|l|l|l|}
\hline
                   & \rule{0pt}{5.5mm}\textbf{LEN-DB}\rule[-2.5mm]{0pt}{2.5mm} & \textbf{SPEECH} & \textbf{EMG} & \textbf{S\&P 500} \\ \hline
\rule{0pt}{5.5mm}\textbf{ConvNetQuake INGV}\rule[-2.5mm]{0pt}{2.5mm}  & 4.79\%          & -0.2\%          & 3.87\%       & 90.44\%           \\ \hline
\rule{0pt}{5.5mm}\textbf{MagNet}\rule[-2.5mm]{0pt}{2.5mm}    & 1.11\%          & -0.3\%          & 1.8\%        & 96.14\%           \\ \hline
\rule{0pt}{5.5mm}\textbf{MLSTM FCN}\rule[-2.5mm]{0pt}{2.5mm} & -3.66\%         & 1.36\%          & 0.34\%       & 95.17\%           \\ \hline
\rule{0pt}{5.5mm}\textbf{TCN}\rule[-2.5mm]{0pt}{2.5mm}       & -0.07\%         & 0.28\%          & 8.29\%       & 96.05\%           \\ \hline
\rule{0pt}{5.5mm}\textbf{Average}\rule[-2.5mm]{0pt}{2.5mm}   & 0.54\%          & 0.29\%          & 3.58\%       & 94.45\%           \\ \hline
\end{tabular}
\end{table}

\subsection{General usefulness of transfer learning}
\label{usefulness_of_transfer_learning}
The main goal of this paper was to investigate whether TL could be useful for TS data in general, by providing a single yes/no answer for each given experimental setup. This can be achieved by counting wins and losses of TL models like it was explained earlier in section~\ref{statistical_tests}. Because the experiment was repeated seven times, and also a grid search for optimal learning rate multiplier value was performed each time (trying ten different multipliers), extra steps must be taken before proceeding with the statistical tests. This results in 70 trained models for each architecture-source-target triplet. Because the goal of rerunning the entire experiment was to obtain more precise statistical metrics, whereas the goal of grid search was to find optimal values, a single value can be computed for each triplet by computing the average score across all runs where for each run only the best achieved score from grid search is taken into account. This is done for all considered triplets. In total, 240 values are obtained in this way. To clarify, this value is obtained by examining all possible cases for the six source domains, six target domains, two variants for each target domain and four different ML architectures. The reader should note that the cases in which source and target datasets are the same are not considered. Rerunning the entire experiment seven times also means that for each target dataset there are seven referent models. So for each case, a referent score value is computed by averaging score values across seven reruns of the same referent model.

It is now possible to compare the obtained TL scores with referent scores, count wins and loses and apply a sign test. Table~\ref{tab:wins_loses_performance} shows the number of wins and losses for intra-domain TL (in which the source and the target domain are seismology) and cross-domain TL which is broken down into three subcases for a more informative view. 

Often, TL can lead to shorter training times which can be beneficial when the used ML model architectures require a lot of time to train. For this reason, we look at the convergence rate of the models from Table~\ref{tab:wins_loses_convergence} to get a general view of how TL affects their convergence. This table was obtained in the same fashion as Table~\ref{tab:wins_loses_performance}, with the exception that for each case the average convergence rate is computed instead of the average performance score. We should note that the best model from grid search is still picked by its performance score and not by its convergence rate. This is because the primary goal is to optimise model performance, and the second goal is to speed up the training process. Therefore, the same models that were compared by their performance scores in Table~\ref{tab:wins_loses_performance}, are now compared by their convergence rate in Table~\ref{tab:wins_loses_convergence}.

\begin{table}[H]
\caption{Comparison of TL models and the corresponding referent models\\ by their performance in terms of wins and loses.}
\label{tab:wins_loses_performance}
\begin{tabular}{|l|l|l|l|}
\hline
                                                                                                            & \rule{0pt}{5.5mm}\textbf{TL wins}\rule[-2.5mm]{0pt}{2.5mm}                                              & \textbf{TL loses}                                            & \textbf{Number of cases} \\ \hline
\textbf{\begin{tabular}[c]{@{}l@{}}\rule{0pt}{5.5mm}Intra-domain\\ (TL between different\\ seismology)\rule[-2.5mm]{0pt}{2.5mm}\end{tabular}} & \textbf{\begin{tabular}[c]{@{}l@{}}43\\ (90\%)\end{tabular}}  & \textbf{\begin{tabular}[c]{@{}l@{}}5\\ (10\%)\end{tabular}}  & \textbf{48}              \\ \hline
\textbf{Cross-domain}                                                                                       & \textbf{\begin{tabular}[c]{@{}l@{}}\rule{0pt}{5.5mm}161\\ (84\%)\rule[-2.5mm]{0pt}{2.5mm}\end{tabular}} & \textbf{\begin{tabular}[c]{@{}l@{}}31\\ (16\%)\end{tabular}} & \textbf{192}             \\ \hline
\begin{tabular}[c]{@{}l@{}}\hspace{5.5mm}\rule{0pt}{5.5mm}TL from\\\hspace{5.5mm}seismology to\\\hspace{5.5mm}other domains\rule[-2.5mm]{0pt}{2.5mm}\end{tabular}                      & \begin{tabular}[c]{@{}l@{}}66\\ (92\%)\end{tabular}           & \begin{tabular}[c]{@{}l@{}}6\\ (8\%)\end{tabular}            & 72                       \\ \hline
\begin{tabular}[c]{@{}l@{}}\hspace{5.5mm}\rule{0pt}{5.5mm}TL from\\\hspace{5.5mm}other domains to the\\\hspace{5.5mm}seismology\rule[-2.5mm]{0pt}{2.5mm}\end{tabular}                  & \begin{tabular}[c]{@{}l@{}}52\\ (72\%)\end{tabular}           & \begin{tabular}[c]{@{}l@{}}20\\ (28\%)\end{tabular}          & 72                       \\ \hline
\begin{tabular}[c]{@{}l@{}}\hspace{5.5mm}\rule{0pt}{5.5mm}TL\\\hspace{5.5mm}between other domains\\\hspace{5.5mm}(SPEECH, EMG, S\&P 500)\rule[-2.5mm]{0pt}{2.5mm}\end{tabular}                & \begin{tabular}[c]{@{}l@{}}43\\ (90\%)\end{tabular}           & \begin{tabular}[c]{@{}l@{}}5\\ (10\%)\end{tabular}           & 48                       \\ \hline
\textbf{Total}                                                                                              & \textbf{\begin{tabular}[c]{@{}l@{}}\rule{0pt}{5.5mm}204\\ (76\%)\rule[-2.5mm]{0pt}{2.5mm}\end{tabular}} & \textbf{\begin{tabular}[c]{@{}l@{}}36\\ (24\%)\end{tabular}} & \textbf{240}             \\ \hline
\end{tabular}
\end{table}
\begin{table}[H]
\caption{Comparison of TL models and the corresponding referent models\\ by their convergence rate in terms of wins and loses.}
\label{tab:wins_loses_convergence}
\begin{tabular}{|l|l|l|l|}
\hline
                                                                                                            & \rule{0pt}{5.5mm}\textbf{TL wins}\rule[-2.5mm]{0pt}{2.5mm}                                              & \textbf{TL loses}                                            & \textbf{Number of cases} \\ \hline
\textbf{\begin{tabular}[c]{@{}l@{}}\rule{0pt}{5.5mm}Intra-domain\\ (TL between different\\ seismology)\rule[-2.5mm]{0pt}{2.5mm}\end{tabular}} & \textbf{\begin{tabular}[c]{@{}l@{}}28\\ (58\%)\end{tabular}}  & \textbf{\begin{tabular}[c]{@{}l@{}}20\\ (42\%)\end{tabular}} & \textbf{48}              \\ \hline
\textbf{Cross-domain}                                                                                       & \textbf{\begin{tabular}[c]{@{}l@{}}\rule{0pt}{5.5mm}125\\ (65\%)\rule[-2.5mm]{0pt}{2.5mm}\end{tabular}} & \textbf{\begin{tabular}[c]{@{}l@{}}67\\ (35\%)\end{tabular}} & \textbf{192}             \\ \hline
\begin{tabular}[c]{@{}l@{}}\hspace{5.5mm}\rule{0pt}{5.5mm}TL from\\\hspace{5.5mm}seismology to\\\hspace{5.5mm}other domains\rule[-2.5mm]{0pt}{2.5mm}\end{tabular}                      & \begin{tabular}[c]{@{}l@{}}52\\ (72\%)\end{tabular}           & \begin{tabular}[c]{@{}l@{}}20\\ (28\%)\end{tabular}          & 72                       \\ \hline
\begin{tabular}[c]{@{}l@{}}\hspace{5.5mm}\rule{0pt}{5.5mm}TL from\\\hspace{5.5mm}other domains to the\\\hspace{5.5mm}seismology\rule[-2.5mm]{0pt}{2.5mm}\end{tabular}                  & \begin{tabular}[c]{@{}l@{}}36\\ (50\%)\end{tabular}           & \begin{tabular}[c]{@{}l@{}}36\\ (50\%)\end{tabular}          & 72                       \\ \hline
\begin{tabular}[c]{@{}l@{}}\hspace{5.5mm}\rule{0pt}{5.5mm}TL\\\hspace{5.5mm}between other domains\\\hspace{5.5mm}(SPEECH, EMG, S\&P 500)\rule[-2.5mm]{0pt}{2.5mm}\end{tabular}                & \begin{tabular}[c]{@{}l@{}}37\\ (77\%)\end{tabular}           & \begin{tabular}[c]{@{}l@{}}11\\ (23\%)\end{tabular}          & 48                       \\ \hline
\textbf{Total}                                                                                              & \textbf{\begin{tabular}[c]{@{}l@{}}\rule{0pt}{5.5mm}153\\ (64\%)\rule[-2.5mm]{0pt}{2.5mm}\end{tabular}} & \textbf{\begin{tabular}[c]{@{}l@{}}87\\ (36\%)\end{tabular}} & \textbf{240}             \\ \hline
\end{tabular}
\end{table}
\begin{table}[H]
\caption{The relationship between the change in convergence rate and performance score.}
\label{tab:relationship_convergence_performance}
\begin{tabular}{|l|c|l|l|}
\hline
\multicolumn{2}{|l|}{\multirow{2}{*}{}}                                                                                                    & \multicolumn{2}{c|}{\rule{0pt}{5.5mm}\textbf{Performance score}\rule[-2.5mm]{0pt}{2.5mm}}                                                                                                                                            \\ \cline{3-4} 
\multicolumn{2}{|l|}{}                                                                                                                     & \multicolumn{1}{c|}{\textbf{\begin{tabular}[c]{@{}c@{}}\rule{0pt}{5.5mm}TL models\\ are worse\rule[-2.5mm]{0pt}{2.5mm}\end{tabular}}} & \multicolumn{1}{c|}{\textbf{\begin{tabular}[c]{@{}c@{}}TL models\\ are better\end{tabular}}} \\ \hline
\multicolumn{1}{|c|}{\multirow{2}{*}{\textbf{Convergence rate}}} & \textbf{\begin{tabular}[c]{@{}c@{}}\rule{0pt}{5.5mm}TL models\\ are worse\rule[-2.5mm]{0pt}{2.5mm}\end{tabular}}  & 10 (4\%)                                                                                    & 77 (32\%)                                                                                    \\ \cline{2-4} 
\multicolumn{1}{|c|}{}                                           & \textbf{\begin{tabular}[c]{@{}c@{}}\rule{0pt}{5.5mm}TL models\\ are better\rule[-2.5mm]{0pt}{2.5mm}\end{tabular}} & 13 (5\%)                                                                                    & 140 (59\%)                                                                                   \\ \hline
\end{tabular}
\end{table} 

The binomial test is applied to the data in Tables~\ref{tab:wins_loses_performance} and \ref{tab:wins_loses_convergence}. The critical value of 136 wins can be obtained using the formula for the sign test, presented in section~\ref{statistical_tests}, where N=240. Because TL models outperformed the referent models in 204 cases, which is shown in Table~\ref{tab:wins_loses_performance}, we can conclude that TL is significantly better than training from scratch for the given experimental setup.

From Table~\ref{tab:wins_loses_performance}, it is clear that intra-domain TL within seismology  yields better results than cross-domain transfer to seismology (90\% vs. 72\%). This makes sense because subdomains within seismology are more related to each other than they are to the other domains. When other domains were used as a target, seismology was equally valuable as a source domain as the other domains (92\% vs. 90\%). This suggests that features learned from seismology are useful for solving problems in other domains we tested. However, TL was not as beneficial when features learned on other domains were transferred to seismology.

Asymmetry can also be seen by looking at the cases of cross-domain TL involving seismology in Table~\ref{tab:wins_loses_performance}. 92\% of TL models outperformed the referent models in the case where seismology was the source domain. In the opposite situation, in which the other domains are the source and seismology is the target domain, 72\% of TL models outperformed their referent counterparts. If there was no difference between two compared groups, the data would show approximately the same number of wins for TL and referent models.

When looking at the comparison of convergence rates in Table~\ref{tab:wins_loses_convergence}, the data shows that TL models were better than referent models in 153 cases, which is higher than the critical value of 136. We can conclude that TL has also enabled faster convergence in addition to the better performance scores. The number of wins by convergence rate is smaller than the number of wins by performance score in all cases which leads to a general conclusion that when training by TL, one is more likely to obtain better results than better convergence.

Correlation between performance score and convergence rate can be seen in Table~\ref{tab:relationship_convergence_performance}. The data suggests that convergence rate does not change if TL did not yield a better performance score. However, the convergence rate has approximately twice the chance (140/77 = 1.82) to increase when TL achieves a better performance score.

\subsection{Domains compatibility}
\label{domains_compatibility}
In the previous section, we discussed the general usefulness of TL. In this section, we present a closer look at the performance and convergence rate differences in intra-domain and cross-domain TL. The second goal of our work was to address the compatibility between different pairs of domains. We hypothesize that some source domains will yield higher performance and convergence rate boost compared to some other source domains.

280 TL models were trained, in total, one for each pair of source and target datasets. This number arises from the fact that four different ML architectures were tested with ten different learning rate multipliers (grid search) and the entire experiment was rerun seven times. Similarly, 28 referent models were trained without TL. The chosen statistical test requires the data to be paired. Because grid search was performed to find the optimal learning rate multiplier, only the model achieving the best performance score is kept from each rerun. This results in 28 TL models coming from four different ML architectures and seven reruns. For each source-target domain pair, a difference in performance scores is computed between TL models and referent models. Then a single difference value is obtained by averaging the computed differences.

Tables~\ref{tab:tl_performance_part1} and \ref{tab:tl_performance_part2} are obtained using the described procedure. Table~\ref{tab:tl_performance_part1} shows situations in which the target domain is seismology, while Table~\ref{tab:tl_performance_part2} shows situations in which the target dataset was from the other chosen domains. The difference between performances is represented as a percentage, for easier comparison. A positive percentage difference represents a decrease in MAE for regression models and an increase in the weighted F1 score for classification. A negative percentage difference represents an increase in MAE for regression models and a decrease in the weighted F1 score for classification models. So, a positive difference always denotes a positive impact of TL, while a negative difference denotes a negative impact of TL.

For each pair of domains, a separate Wilcoxon signed-ranks test is performed which leads to a total of 60 statistical tests (30 in each table). Therefore, the significance level of 0.05 requires an adjustment for this problem. For this purpose, we use the Benjamini-Krieger-Yekutieli method to combat the problem of multiple comparisons and adjust significance level accordingly. Colored cells are found to be statistically significant. Green color shows that TL models outperformed referent models (positive transfer), while the red color shows the opposite situation (negative transfer).

\begin{table}[]
\caption{Comparison of performance scores between TL models and referent models with seismology being the target domain.}
\label{tab:tl_performance_part1}
\begin{adjustbox}{width=\columnwidth,center}
\begin{tabular}{|l|l|l|l|l|l|l|}
\hline
                   & \multicolumn{6}{c|}{\rule{0pt}{5.5mm}\textbf{Target datasets}\rule[-2.5mm]{0pt}{2.5mm}}                                                                                                                                                                                                               \\ \cline{2-7} 
\multirow{-2}{*}{} & \rule{0pt}{5.5mm}\textbf{LOMAX 1k5}\rule[-2.5mm]{0pt}{2.5mm}                      & \textbf{LOMAX 9k}                       & \textbf{LEN-DB 1k5}                     & \textbf{LEN-DB 9k}                      & \textbf{STEAD 1k5}                       & \textbf{STEAD 9k}                        \\ \hline
\rule{0pt}{5.5mm}\textbf{LOMAX}\rule[-2.5mm]{0pt}{2.5mm}\rule[-2.5mm]{0pt}{2.5mm}     & -                                       & -                                       & \cellcolor[HTML]{34FF34}\textbf{8.21\%} & \cellcolor[HTML]{34FF34}\textbf{3.17\%} & \cellcolor[HTML]{34FF34}\textbf{6.63\%}  & \cellcolor[HTML]{34FF34}\textbf{3.96\%}  \\ \hline
\rule{0pt}{5.5mm}\textbf{LEN-DB}\rule[-2.5mm]{0pt}{2.5mm}    & \cellcolor[HTML]{34FF34}\textbf{8.25\%} & \cellcolor[HTML]{34FF34}\textbf{4.01\%} & -                                       & -                                       & \cellcolor[HTML]{34FF34}\textbf{2.49\%}  & \cellcolor[HTML]{34FF34}\textbf{2.52\%}  \\ \hline
\rule{0pt}{5.5mm}\textbf{STEAD}\rule[-2.5mm]{0pt}{2.5mm}     & 1.16\%                                  & -1.87\%                                 & 0.56\%                                  & -0.52\%                                 & -                                        & -                                        \\ \hline
\rule{0pt}{5.5mm}\textbf{SPEECH}\rule[-2.5mm]{0pt}{2.5mm}    & \cellcolor[HTML]{34FF34}\textbf{9.29\%} & \cellcolor[HTML]{34FF34}\textbf{3.58\%} & \cellcolor[HTML]{34FF34}\textbf{4.47\%} & \cellcolor[HTML]{34FF34}\textbf{3.19\%} & \cellcolor[HTML]{FE0000}\textbf{-2.31\%} & 0.07\%                                   \\ \hline
\rule{0pt}{5.5mm}\textbf{EMG}\rule[-2.5mm]{0pt}{2.5mm}       & \cellcolor[HTML]{34FF34}\textbf{9.12\%} & \cellcolor[HTML]{34FF34}\textbf{5.34\%} & -0.06\%                                 & 0.41\%                                  & 1.15\%                                   & \cellcolor[HTML]{34FF34}\textbf{2.04\%}  \\ \hline
\rule{0pt}{5.5mm}\textbf{S\&P 500}\rule[-2.5mm]{0pt}{2.5mm}  & 0.12\%                                  & 0.12\%                                  & -0.92\%                                 & 1.14\%                                  & \cellcolor[HTML]{FE0000}\textbf{-7.23\%} & \cellcolor[HTML]{FE0000}\textbf{-2.29\%} \\ \hline
\end{tabular}
\end{adjustbox}
\end{table}
\begin{table}[]
\caption{Comparison of performance scores between TL models and referent models when the target dataset is SPEECH, EMG or S\&P 500.}
\label{tab:tl_performance_part2}
\begin{adjustbox}{width=\columnwidth,center}
\begin{tabular}{|l|l|l|l|l|l|l|}
\hline
                   & \multicolumn{6}{c|}{\rule{0pt}{5.5mm}\textbf{Target datasets}\rule[-2.5mm]{0pt}{2.5mm}}                                                                                                                                                                                                             \\ \cline{2-7} 
\multirow{-2}{*}{} & \rule{0pt}{5.5mm}\textbf{SPEECH 1k5}\rule[-2.5mm]{0pt}{2.5mm}                     & \textbf{SPEECH 9k}                      & \textbf{EMG 1k5}                        & \textbf{EMG 9k}                         & \textbf{S\&P 500 1k5}                   & \textbf{S\&P 500 9k}                    \\ \hline
\rule{0pt}{5.5mm}\textbf{LOMAX}\rule[-2.5mm]{0pt}{2.5mm}     & \cellcolor[HTML]{34FF34}\textbf{45.7\%} & \cellcolor[HTML]{34FF34}\textbf{871\%}  & \cellcolor[HTML]{34FF34}\textbf{24.2\%} & \cellcolor[HTML]{34FF34}\textbf{17.6\%} & \cellcolor[HTML]{34FF34}\textbf{45.4\%} & \cellcolor[HTML]{34FF34}\textbf{22.8\%} \\ \hline
\rule{0pt}{5.5mm}\textbf{LEN-DB}\rule[-2.5mm]{0pt}{2.5mm}    & \cellcolor[HTML]{34FF34}\textbf{309\%}  & \cellcolor[HTML]{34FF34}\textbf{1210\%} & \cellcolor[HTML]{34FF34}\textbf{21.9\%} & \cellcolor[HTML]{34FF34}\textbf{12.1\%} & \cellcolor[HTML]{34FF34}\textbf{44.4\%} & \cellcolor[HTML]{34FF34}\textbf{22.1\%} \\ \hline
\rule{0pt}{5.5mm}\textbf{STEAD}\rule[-2.5mm]{0pt}{2.5mm}     & \cellcolor[HTML]{34FF34}\textbf{323\%}  & \cellcolor[HTML]{34FF34}\textbf{1080\%} & \cellcolor[HTML]{34FF34}\textbf{18.7\%} & \cellcolor[HTML]{34FF34}\textbf{10\%}   & \cellcolor[HTML]{34FF34}\textbf{44.3\%} & \cellcolor[HTML]{34FF34}\textbf{22.7\%} \\ \hline
\rule{0pt}{5.5mm}\textbf{SPEECH}\rule[-2.5mm]{0pt}{2.5mm}    & \textbf{-}                              & \textbf{-}                              & \cellcolor[HTML]{34FF34}\textbf{13.5\%} & 9.87\%                                  & \cellcolor[HTML]{34FF34}\textbf{44.1\%} & \cellcolor[HTML]{34FF34}\textbf{21.6\%} \\ \hline
\rule{0pt}{5.5mm}\textbf{EMG}\rule[-2.5mm]{0pt}{2.5mm}       & \cellcolor[HTML]{34FF34}\textbf{367\%}  & \cellcolor[HTML]{34FF34}\textbf{1820\%} & -                                       & -                                       & \cellcolor[HTML]{34FF34}\textbf{41.8\%} & \cellcolor[HTML]{34FF34}\textbf{18.2\%} \\ \hline
\rule{0pt}{5.5mm}\textbf{S\&P 500}\rule[-2.5mm]{0pt}{2.5mm}  & \cellcolor[HTML]{34FF34}\textbf{122\%}  & \cellcolor[HTML]{34FF34}\textbf{885\%}  & \cellcolor[HTML]{34FF34}\textbf{15.2\%} & \cellcolor[HTML]{34FF34}\textbf{11.9\%} & \textbf{-}                              & \textbf{-}                              \\ \hline
\end{tabular}
\end{adjustbox}
\end{table}

The data in Table~\ref{tab:tl_performance_part1} shows that it was beneficial to transfer features learned on the LOMAX and LEN-DB datasets to other seismological datasets. The same cannot be stated for the STEAD dataset because models pre-trained on this dataset had no statistically significant differences compared to the models trained from scratch when fine-tuned on the other two seismological datasets. Because LEN-DB and STEAD datasets contain local earthquakes, the difference being that instrument response is removed from the LEN-DB while it is not removed from STEAD waveforms, and the sampling rates of the data being 20 Hz and 100 Hz for LEN-DB and STEAD, respectively, we speculate that this may be the cause why the transfer of knowledge from LEN-DB to STEAD is useful, while it is not useful in the other direction. The LOMAX dataset, being the dataset of earthquakes recorded at any distance, also has the instrument response removed from the waveforms and the data are sampled with a sampling rate of 20 Hz, which may explain the successful transfer between LOMAX and LEN-DB in both directions and a one-way transfer from LOMAX to STEAD. Another important difference between STEAD and the two other seismological datasets is that the \textit{stream max} input was not used when pre-training and training the models on STEAD.

Knowledge transfer from non-seismological domains to STEAD and LEN-DB datasets does not have the same impact on the results despite their similarities. Models pre-trained on EMG did not have any statistically significant improvement for LEN-DB while they were useful for STEAD 9k. However, models pre-trained on SPEECH were an adequate choice for LEN-DB while they were a bad option for STEAD 1.5k because of the negative transfer.

One can see from Tables~\ref{tab:tl_performance_part1} and \ref{tab:tl_performance_part2} that negative transfer was only observed when the source domain was either SPEECH or S\&P 500 and the target domain was STEAD. In the case of SPEECH as a source domain, we found out that negative transfer occurred with all the models except the MagNet. In the case of S\&P 500 source domain, the same was observed for the 1k5 variant, while on the 9k variant MLSTM FCN was also not affected anymore along with the MagNet. Both of these examples show how the effects of negative transfer are dropping (getting closer to zero) as the target training dataset size gets larger.

When looking at the case of SPEECH as the target dataset in Table~\ref{tab:tl_performance_part2}, one can see an unusually high increase in performance. For this reason we investigated this case more closely. We noticed that all referent models except the MLSTM FCN did not converge, both on 1k5 and on 9k variants. TL was found to be beneficial to all models regardless of the source domains in these cases. This is true even for the models pre-trained on S\&P 500 dataset which may be considered mostly “random” and almost “impossible to predict”. The results suggest that the SPEECH dataset is very challenging because of a small number of training instances, and TL is of great help in this case.

In the case of EMG target dataset, the table shows how the models pre-trained on SPEECH were a good choice for the 1k5 variant, while in the case of the 9k variant these models achieved the same results as the reference models. In contrast, models pre-trained on other datasets performed better than the referent ones on both 1k5 and 9k variants. This does suggest that some source domains are useful even when a larger quantity of training instances are available while some other domains are not. Along with the SPEECH, this was the only target domain that benefited from \textit{a priori} gained knowledge on S\&P 500 dataset.

In the case of S\&P 500 as a target dataset, it seems like the TL boost remains almost the same regardless of the source dataset. We examined this case closer. We found out that MAE of ConvNetQuake INGV and MLSTM FCN models was approximately five times higher than that of the other models. The same problem persisted only for the ConvNetQuake INGV in the case of the 9k variant. This is the reason why the boost for the 9k variant halved with respect to the 1k5 variant. TL helped those two models to achieve roughly equal MAE as all other models.

It is logical to expect that the effects of TL will become less noticeable as the size of the target training set increases. The reasoning behind this is that as the target dataset increases, it provides more and more training instances and this leads to a better performance score. This automatically reduces the need for TL as the target dataset alone contains enough data for successful training. That seems to be true in most cases. For example, in the case when the source dataset is EMG and the target domain is LOMAX, the performance gain is lower on the 9k than on the 1k5 variant. However, there are a few cases where models achieved a greater performance boost on the 9k variant in comparison to the 1k5 variant. Those cases are: EMG $\rightarrow$ STEAD and LEN-DB $\rightarrow$ STEAD.

We examined those cases more closely. We hypothesise this may be due to the STEAD waveforms containing the instrument response which makes the task more difficult to learn. This would suggest that the usefulness of TL depends on the size of the target training set. When the target training set is extremely small, referent and TL models perform equally well and there is no benefit of TL because there are not enough training instances to learn a meaningful representation. As the size of the target training set increases, the difference in performance score between the TL models and the referent models becomes higher. At some point, the target training set contains enough data for successful training from scratch. From that point on, the benefits of TL start to vanish as the target training set size continues to grow. The same is true in the case of negative transfer when models perform worse than the referent ones. As it was already explained, these negative effects also wear off as the size of the target training set continues to grow. Of course, these “boundary” sizes are different for each domain (i.e. dependent on the complexity of the task) and ML architecture used which can explain why referent and TL models perform equally well on some target datasets, while on the other domains they perform differently for the same target dataset size.

\begin{table}[]
\caption{Comparison of convergence rate between TL models and referent models with seismology being the target domain.}
\label{tab:tl_convergence_part1}
\begin{adjustbox}{width=\columnwidth,center}
\begin{tabular}{|l|l|l|l|l|l|l|}
\hline
                   & \multicolumn{6}{c|}{\rule{0pt}{5.5mm}\textbf{Target datasets}\rule[-2.5mm]{0pt}{2.5mm}}                                                                                                                                            \\ \cline{2-7} 
\multirow{-2}{*}{} & \rule{0pt}{5.5mm}\textbf{LOMAX 1k5}\rule[-2.5mm]{0pt}{2.5mm}                      & \textbf{LOMAX 9k} & \textbf{LEN-DB 1k5}                     & \textbf{LEN-DB 9k} & \textbf{STEAD 1k5}                      & \textbf{STEAD 9k} \\ \hline
\rule{0pt}{5.5mm}\textbf{LOMAX}\rule[-2.5mm]{0pt}{2.5mm}     & -                                       & -                 & 9.08\%                                  & -3.15\%            & 16.6\%                                  & 11.6\%            \\ \hline
\rule{0pt}{5.5mm}\textbf{LEN-DB}\rule[-2.5mm]{0pt}{2.5mm}    & \cellcolor[HTML]{34FF34}\textbf{11.6\%} & 6.72\%            & -                                       & -                  & 17.5\%                                  & 11.8\%            \\ \hline
\rule{0pt}{5.5mm}\textbf{STEAD}\rule[-2.5mm]{0pt}{2.5mm}     & \cellcolor[HTML]{34FF34}\textbf{16.5\%} & 4.59\%            & \cellcolor[HTML]{34FF34}\textbf{22.6\%} & 7.59\%             & -                                       & -                 \\ \hline
\rule{0pt}{5.5mm}\textbf{SPEECH}\rule[-2.5mm]{0pt}{2.5mm}    & 10.2\%                                  & 3.19\%            & \cellcolor[HTML]{34FF34}\textbf{5.45\%} & 1.35\%             & 6.73\%                                  & 8.71\%            \\ \hline
\rule{0pt}{5.5mm}\textbf{EMG}\rule[-2.5mm]{0pt}{2.5mm}       & 10.1\%                                  & 5.76\%            & 14.9\%                                  & -0.31\%            & 8.98\%                                  & 12.8\%            \\ \hline
\rule{0pt}{5.5mm}\textbf{S\&P 500}\rule[-2.5mm]{0pt}{2.5mm}  & 4.04\%                                  & 4.33\%            & 13.8\%                                  & -1.39\%            & \cellcolor[HTML]{34FF34}\textbf{9.76\%} & 13.9\%            \\ \hline
\end{tabular}
\end{adjustbox}
\end{table}
\begin{table}[]
\caption{Comparison of convergence rate between TL models and referent models when the target dataset is SPEECH, EMG or S\&P 500.}
\label{tab:tl_convergence_part2}
\begin{adjustbox}{width=\columnwidth,center}
\begin{tabular}{|l|l|l|l|l|l|l|}
\hline
                   & \multicolumn{6}{c|}{\rule{0pt}{5.5mm}\textbf{Target datasets}\rule[-2.5mm]{0pt}{2.5mm}}                                                                                                                                             \\ \cline{2-7} 
\multirow{-2}{*}{} & \rule{0pt}{5.5mm}\textbf{SPEECH 1k5}\rule[-2.5mm]{0pt}{2.5mm} & \textbf{SPEECH 9k}             & \textbf{EMG 1k5}               & \textbf{EMG 9k}                 & \textbf{S\&P 500 1k5}          & \textbf{S\&P 500 9k}           \\ \hline
\rule{0pt}{5.5mm}\textbf{LOMAX}\rule[-2.5mm]{0pt}{2.5mm}     & 0.12\%              & 4.55\%                         & \cellcolor[HTML]{34FF34}13.3\% & -1.89\%                         & \cellcolor[HTML]{34FF34}6.37\% & \cellcolor[HTML]{34FF34}5.39\% \\ \hline
\rule{0pt}{5.5mm}\textbf{LEN-DB}\rule[-2.5mm]{0pt}{2.5mm}    & 1.13\%              & \cellcolor[HTML]{34FF34}7.87\% & \cellcolor[HTML]{34FF34}17.0\% & -3.67\%                         & \cellcolor[HTML]{34FF34}6.43\% & 3.66\%                         \\ \hline
\rule{0pt}{5.5mm}\textbf{STEAD}\rule[-2.5mm]{0pt}{2.5mm}     & 2.35\%              & 3.4\%                          & \cellcolor[HTML]{34FF34}17.3\% & \cellcolor[HTML]{FE0000}-7.19\% & \cellcolor[HTML]{34FF34}6.16\% & \cellcolor[HTML]{34FF34}9.7\%  \\ \hline
\rule{0pt}{5.5mm}\textbf{SPEECH}\rule[-2.5mm]{0pt}{2.5mm}    & \textbf{-}          & \textbf{-}                     & \cellcolor[HTML]{34FF34}13.9\% & -5.11\%                         & \cellcolor[HTML]{34FF34}6.13\% & \cellcolor[HTML]{34FF34}3.13\% \\ \hline
\rule{0pt}{5.5mm}\textbf{EMG}\rule[-2.5mm]{0pt}{2.5mm}       & 5.76\%              & \cellcolor[HTML]{34FF34}9.35\% & -                              & -                               & \cellcolor[HTML]{34FF34}6.2\%  & \cellcolor[HTML]{34FF34}8.25\% \\ \hline
\rule{0pt}{5.5mm}\textbf{S\&P 500}\rule[-2.5mm]{0pt}{2.5mm}  & 4.8\%               & 5.75\%                         & 8.12\%                         & \cellcolor[HTML]{FE0000}-6.09\% & \textbf{-}                     & \textbf{-}                     \\ \hline
\end{tabular}
\end{adjustbox}
\end{table}

Tables~\ref{tab:tl_convergence_part1} and \ref{tab:tl_convergence_part2} show the change in convergence rate for each pair of TL and referent models that were examined in Tables~\ref{tab:tl_convergence_part1} and \ref{tab:tl_convergence_part2}. A difference in their convergence rate is computed and these differences are averaged and expressed in percentage with respect to the average convergence rate of referent models. In this way, a single value for each cell in the table is obtained. Statistical tests and Benjamini-Krieger-Yekutieli correction are performed in the same way as for Tables~\ref{tab:tl_convergence_part1} and \ref{tab:tl_convergence_part2}.

Table~\ref{tab:tl_convergence_part1} shows that the knowledge gained on the LOMAX and the LEN-DB datasets did not statistically significantly accelerate the convergence rate on the STEAD target tasks. On the other hand, the knowledge gained on STEAD accelerated convergence on LOMAX and LEN-DB target tasks. Also, the table shows that the fine-tuned models on LOMAX and LEN-DB achieved faster convergence by knowledge transfer in more cases compared to STEAD. This is in opposition with the observations made by performance score for the same cases. This is an example that demonstrates that a boost in performance does not necessarily lead to a boost in convergence rate as was concluded in section~\ref{usefulness_of_transfer_learning}.

The data shows that almost all models had no statistically significant differences in convergence compared to the referent models in the case of SPEECH target tasks. However, this comparison is not so informative since we previously determined that those referent models did not learn any meaningful representation (Table~\ref{tab:tl_performance_part1}).

EMG target datasets are the only ones for whom we observed a significantly slower convergence of TL models in comparison to the referent models on the greater dataset, while the convergence rate was significantly faster on the smaller one. Upon inspecting this case closer, we found out that all four architectures exhibit this behaviour. The reason for such behaviour is unclear.

In the case of S\&P 500 target dataset, we found that convergence of ConvNetQuake INGV, MagNet and MLSTM FCN models was sped up by the TL. For each model, this speedup was approximately the same across all source domains. This explains why it looks like the speedup is constant regardless of the source domain.

The readers are referred to the supplementary materials to get more detailed results.

\subsection{The influence of models on TL}
\label{influence_of_models_on_tl}

\begin{table}[H]
\caption{The influence of the chosen model on the TL performance gain. The numbers\\ show how many times each TL model outperformed its corresponding referent model.}
\label{tab:influence_of_chosen_model}
\begin{tabular}{|l|l|l|l|l|}
\hline
                                                                                                                 & \textbf{\begin{tabular}[c]{@{}l@{}}\rule{0pt}{5.5mm}ConvNetQuake\\ INGV\rule[-2.5mm]{0pt}{2.5mm}\end{tabular}} & \textbf{MagNet}                                                  & \textbf{MLSTM FCN}                                               & \textbf{TCN}                                                    \\ \hline
\textbf{\begin{tabular}[c]{@{}l@{}}\rule{0pt}{5.5mm}Intra-domain\\ (TL between different\\ seismology datasets)\rule[-2.5mm]{0pt}{2.5mm}\end{tabular}}     & \textbf{\begin{tabular}[c]{@{}l@{}}75\%\\ (9/12)\end{tabular}}       & \textbf{\begin{tabular}[c]{@{}l@{}}100\%\\ (12/12)\end{tabular}} & \textbf{\begin{tabular}[c]{@{}l@{}}100\%\\ (12/12)\end{tabular}} & \textbf{\begin{tabular}[c]{@{}l@{}}83\%\\ (10/12)\end{tabular}} \\ \hline
\textbf{Cross-domain}                                                                                            & \textbf{\begin{tabular}[c]{@{}l@{}}\rule{0pt}{5.5mm}88\%\\ (42/48)\rule[-2.5mm]{0pt}{2.5mm}\end{tabular}}      & \textbf{\begin{tabular}[c]{@{}l@{}}88\%\\ (42/48)\end{tabular}}  & \textbf{\begin{tabular}[c]{@{}l@{}}81\%\\ (39/48)\end{tabular}}  & \textbf{\begin{tabular}[c]{@{}l@{}}79\%\\ (38/48)\end{tabular}} \\ \hline
\begin{tabular}[c]{@{}l@{}}\hspace{5.5mm}\rule{0pt}{5.5mm}TL from\\\hspace{5.5mm}seismology to\\\hspace{5.5mm}other domains\rule[-2.5mm]{0pt}{2.5mm}\end{tabular}               & \begin{tabular}[c]{@{}l@{}}94\%\\ (17/18)\end{tabular}               & \begin{tabular}[c]{@{}l@{}}100\%\\ (18/18)\end{tabular}          & \begin{tabular}[c]{@{}l@{}}78\%\\ (14/18)\end{tabular}           & \begin{tabular}[c]{@{}l@{}}94\%\\ (17/18)\end{tabular}          \\ \hline
\begin{tabular}[c]{@{}l@{}}\hspace{5.5mm}\rule{0pt}{5.5mm}TL from\\\hspace{5.5mm}other domain to\\\hspace{5.5mm}seismology\rule[-2.5mm]{0pt}{2.5mm}\end{tabular}                & \begin{tabular}[c]{@{}l@{}}72\%\\ (13/18)\end{tabular}               & \begin{tabular}[c]{@{}l@{}}72\%\\ (13/18)\end{tabular}           & \begin{tabular}[c]{@{}l@{}}83\%\\ (15/18)\end{tabular}           & \begin{tabular}[c]{@{}l@{}}61\%\\ (11/18)\end{tabular}          \\ \hline
\begin{tabular}[c]{@{}l@{}}\hspace{5.5mm}\rule{0pt}{5.5mm}TL\\\hspace{5.5mm}between other domains\\\hspace{5.5mm}(SPEECH, EMG, S\&P\\\hspace{5.5mm}500)\rule[-2.5mm]{0pt}{2.5mm}\end{tabular} & \begin{tabular}[c]{@{}l@{}}100\%\\ (12/12)\end{tabular}              & \begin{tabular}[c]{@{}l@{}}92\%\\ (11/12)\end{tabular}           & \begin{tabular}[c]{@{}l@{}}83\%\\ (10/12)\end{tabular}           & \begin{tabular}[c]{@{}l@{}}83\%\\ (10/12)\end{tabular}          \\ \hline
\textbf{Total}                                                                                                   & \textbf{\begin{tabular}[c]{@{}l@{}}\rule{0pt}{5.5mm}85\%\\ (51/60)\rule[-2.5mm]{0pt}{2.5mm}\end{tabular}}      & \textbf{\begin{tabular}[c]{@{}l@{}}90\%\\ (54/60)\end{tabular}}  & \textbf{\begin{tabular}[c]{@{}l@{}}85\%\\ (51/60)\end{tabular}}  & \textbf{\begin{tabular}[c]{@{}l@{}}80\%\\ (48/60)\end{tabular}} \\ \hline
\end{tabular}
\end{table}

\begin{figure}[htbp]
    \centering
    \includegraphics[width=11cm, keepaspectratio]{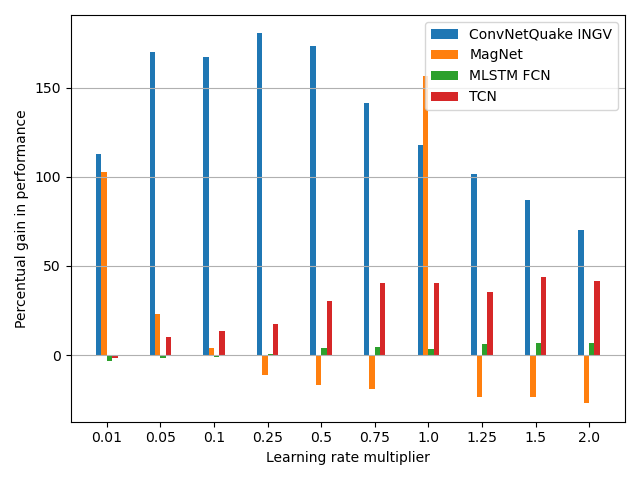}
    \caption{Impact of learning rate multiplier on the TL performance gain. The performance gain is computed for each model and learning rate multiplier separately. It represents the average increase of predictive performance of TL models in comparison to the referent models over all pairs of source and target datasets.}
    \label{fig:learning_rate_impact}
\end{figure}

Table~\ref{tab:influence_of_chosen_model} shows how many times each TL model outperformed its corresponding referent model. It is important to keep in mind that this table does not reflect performance score, but rather to what extent was \textit{a priori} gained knowledge utilised by the models.

The data from the table shows that the results of TL can significantly vary with the chosen model. For example, MagNet and MLSTM FCN were found to make the best use of pre-learned knowledge in intra-domain scenarios, while ConvNetQuake INGV and TCN performed significantly worse. However, MagNet's, MLSTM FCN's and TCN's success of utilising TL dropped significantly in cross-domain transfer to seismology. The reason for the drop is the previously determined lower compatibility of knowledge when transferring from other domains to seismology. Meanwhile, ConvNetQuake INGV performed almost the same as in the intra-domain scenario. Even if ConvNetQuake INGV and MagNet were both designed primarily for the domain of seismology, they exhibit different behavior during intra and cross-domain knowledge transfer to seismology. The same can be said for the general purpose models: MLSTM FCN was found to be a better option when the target domain was seismology, while in the other two cases they performed equally well or TCN performed better.

In total, all models utilised a priori gained knowledge in 80\%-90\% of the examined cases. Therefore, one could expect to get some of the benefits of TL with a preselected model, but must be aware of the fact that better performances may be gained by some other model.
	
To which extent is a priori gained knowledge useful also depends on the hyperparameters of TL. In our case, we inspected ten different values for the learning rate multiplier hyperparameter - through grid search - and present the results in Figure~\ref{fig:learning_rate_impact}. The figure illustrates how different learning rate multipliers affected models performance when training with TL, expressed as percentage. A positive percentage means that TL outperformed the referent models, and vice-versa. It can be seen that different multiplier values have noticeably different effects to which extent the TL approach will be successful. ConvNetQuake INGV utilises a priori gained knowledge the best when multiplier value is small, and the best performance was achieved when it was 0.25. TCN is similar to the ConvNetQuake INGV in the sense that both of these models consist of a larger number of convolutional layers, but regardless of that similarity, TCN was found to perform best using multipliers 0.75 and above.

MagNet and MLSTM FCN models are similar in sense that both of them have fewer convolutional layers and contain LSTM units. They also exhibit different behaviour regardless of these similarities. MLSTM FCN works best with multiplier values of 0.5 or above. Magnet model was found to achieve the greatest gain when a multiplier of 1.0 or 0.01 was used. It experiences effects of negative transfer with multipliers greater than 0.1, with 1.0 being the exception. This exception makes the behavior of the MagNet model uniquely different from the behavior of other examined models.

Practitioners mostly use learning rate multipliers smaller than 1.0 as an indirect mechanism to preserve the knowledge from the source task and to use it to maximize the performance score on the target task\citep{li2017learning}. However, Figure~\ref{fig:learning_rate_impact} shows that multipliers greater than 1.0 can also be beneficial in cases of some models. This suggests that some models can make use of \textit{a priori} gained knowledge even at learning rate multipliers greater than 1.0 and that knowledge in convolutional layers is not “lost”.

\section{Conclusion}
\label{conclusion}
In this paper, we have broadly explored the effects of TL from TS data to find out how successful TL can solve upcoming problems with the adoption of deep learning models for TS data. As the models become deeper and more complex, so does the need for larger training sets. TL is very well studied in some fields (such as in image classification) and has proven to be one of the good ways to tackle the problem of smaller training sets.

We conducted an experiment through which we examined intra-domain TL situations between different seismological datasets as well as cross-domain TL between seismology, sound, medicine, and financial datasets. We ensured that all selected datasets were of sufficient size for successful pre-training, and we conducted a grid-search to find the optimal hyperparameters for successful knowledge transfer. Finally, we observed the effects of TL on model performance and convergence rate by comparing TL models with those trained from scratch. The models were compared on two target-domain training sets of different sizes, one having 1,500 training instances and the other having 9,000 training instances.

Our experiment showed that there is a statistically significant difference in performance and convergence rate between TL models and models trained from scratch. In 76\% of cases there was an improvement in the performance of the model, while in 64\% of cases there was an acceleration of convergence. The data showed that if the transfer of knowledge from the first to the second domain leads to an improvement, it is not necessary that the transfer of knowledge in the opposite direction will also lead to an improvement. TL models were twice as likely to converge faster than the models trained from scratch if TL led to a better performance.

When we came to high level conclusions, we went to study each pair of source-target domains separately. The SPEECH target dataset proved to be very challenging with small training sets. The vast majority of models trained from scratch failed to learn anything on such small data sets, while TL models managed to converge. This is even true for the transfer of knowledge acquired on the S\&P 500 dataset which can be considered “random” and “unpredictable”. In other cross-domain cases, TL models vastly outperformed the models trained from scratch. In intra-domain TL cases, knowledge has been shown to be transferable between different seismological datasets, but we have also observed that other things could be a factor on which the success of knowledge transfer depends.

Finally, we examined the impact of model selection on the success of knowledge transfer. Our data show that in general all four models outperformed the models trained from scratch in 80\%-90\% of cases. However, in some cases, not all models were equally successful. The success of knowledge transfer is greatly influenced by the hyperparameters related to TL, and in our case it was only the learning rate multiplier that adjusts the fine-tuning process of convolutional layers. We found the optimal value of this hyperparameter for each model in each scenario through grid search; it was not the same for all models, but in most cases it was less than or equal to 1.0. However, in some cases and for some models, values greater than 1.0 have been shown to function well, which could be examined in terms of differential learning rate research in future work.

We started with the idea that all TS data are essentially signals that all TS data can be decomposed into a linear combination of sine and cosine waves, which could indicate that there is common knowledge that can be used to solve problems in various TS domains. We experimented with six intra-domain and 24 cross-domain scenarios to determine to what extent this assumption holds. We found only two source-target domain pairs that result in TL models performing statistically significantly worse than the referent models (i.e. negative transfer). This indicates that if TL is applied without any prior knowledge about compatibility between source and target domains, one is very likely to get a better performance score, or at least as good as the model trained from scratch. By analyzing the pairs of source and target domains, we determined how even seemingly unrelated domains can be mutually compatible enough to yield positive effects (i.e. positive transfer). Ultimately, we found that all models had approximately the same probability of achieving better results with knowledge transfer. All this leads to the conclusion that pre-training models in the field of TS is useful and should be given a chance even when the two TS tasks seem unrelated. The application of TL for TS is yet to become a popular and powerful choice as it has already \textit{revolutionised} many other fields.

\section{Acknowledgement}
This work was supported in part by Croatian Science Foundation projects UIP-2019-04-7999, DOK-2020-01-4659 and IP-2020-02-3770, and the COST project G2Net CA17137 A network for Gravitational Waves, Geophysics and Machine Learning.

This research was performed using the resources of computer cluster Isabella based in SRCE - University of Zagreb University Computing Centre.

\section{Abbreviations}
\begin{table}[H]
\begin{tabular}{ll}
AUC & area under the curve \\
CNN & convolutional neural network \\
EMG & electromyography \\
GPU & graphical processing unit \\
LSTM & long short-term memory \\
MAE & mean absolute error \\
ML & machine learning \\
MLSTM FCN & multivariate LSTM fully convolutional network \\
MSE & mean squared error \\
RNN & recurrent neural network \\
TCN & temporal convolutional network \\
TL & transfer learning \\
TS & time series
\end{tabular}
\end{table}



 \bibliographystyle{elsarticle-num} 
 \bibliography{cas-refs}





\end{document}